\documentclass{article}

\usepackage[main, final]{neurips_2026}
\usepackage{amsmath}
\usepackage{iftex}

\usepackage[utf8]{inputenc} 
\usepackage[T1]{fontenc}    
\usepackage{hyperref}       
\usepackage{url}            
\usepackage{graphicx}       
\usepackage{capt-of}        
\usepackage{booktabs}       
\usepackage{longtable}      
\usepackage{multirow}       
\usepackage{amsfonts}       
\usepackage{nicefrac}       
\usepackage{microtype}      
\usepackage[table]{xcolor}  

\ifXeTeX
  \usepackage{xeCJK}
\fi

\newcommand{\modelname}{\textsc{\textbf{D}isentangled \textbf{E}FM \textbf{F}ine-\textbf{T}uning}}
\newcommand{\modelnickname}{\textsc{\textbf{DEFT}}}
\newcommand{\mixedtokens}{\mathbf{Z}}
\newcommand{\contenttokens}{\mathbf{C}}
\newcommand{\styletokens}{\mathbf{S}}
\definecolor{mixedsand}{RGB}{178,145,84}
\definecolor{contentwine}{HTML}{C69184}
\definecolor{styleblue}{RGB}{104,142,173}
\newcommand{\mixedprose}{\ensuremath{\textcolor{mixedsand}{\mathbf{Z}}}}
\newcommand{\contentprose}{\ensuremath{\textcolor{contentwine}{\mathbf{C}}}}
\newcommand{\styleprose}{\ensuremath{\textcolor{styleblue}{\mathbf{S}}}}
\newcommand{\observationtitle}[1]{\par\vspace{2pt}\noindent\textbf{#1.}\ }
\newcommand{\notetitle}[1]{\par\vspace{2pt}\noindent\textbf{#1.}\ }
\newcommand{\derivationstart}{\par\vspace{2pt}\noindent\textit{Derivation.}\ }
\graphicspath{{images/}}

\title{\modelnickname{}: Disentanglement-Enhanced Fine-Tuning for EEG Foundation Models}

\author{%
  \textbf{Yuqiu Deng\textsuperscript{1} \quad
  Muyun Jiang\textsuperscript{2} \quad
  Zongpeng Zhang\textsuperscript{3} \quad
  Mingqing Xiao\textsuperscript{4} \quad
  Yansen Wang\textsuperscript{4}} \\
  \textbf{Peiliang Gong\textsuperscript{2} \quad
  Xinliang Zhou\textsuperscript{1} \quad
  Yi Ding\textsuperscript{2} \quad
  Ziyu Jia\textsuperscript{5} \quad
  Cuntai Guan\textsuperscript{2} \quad
  Chenyu Liu\textsuperscript{1}} \\
  \textsuperscript{1}Uploading Inc \\
  \textsuperscript{2}College of Computing and Data Science, Nanyang Technological University \\
  \textsuperscript{3}Department of Biostatistics, School of Public Health, Peking University \\
  \textsuperscript{4}Microsoft Research Asia \quad
  \textsuperscript{5}Institute of Automation, Chinese Academy of Sciences \\
  \texttt{yuqiudeng@uploading.tech, james.jiang@ntu.edu.sg, zongpeng.zhang98@gmail.com} \\
  \texttt{mingqingxiao@microsoft.com, yansenwang@microsoft.com, plgong@outlook.com} \\
  \texttt{xinliang@uploading.tech, ding.yi@ntu.edu.sg, jia.ziyu@outlook.com} \\
  \texttt{ctguan@ntu.edu.sg, chenyuliu@uploading.tech}
}

\begin{document}

\maketitle
\begin{abstract}
EEG foundation models (EFMs) have improved EEG representation learning by transferring knowledge from large-scale multi-subject pretraining to downstream tasks. However, their pretrained embeddings often retain substantial subject-related variation, placing an additional burden on the downstream classifier, especially in cross-subject settings. To address this issue, we propose \modelname{} (\modelnickname{}), a plug-and-play adapter for EFM fine-tuning. \modelnickname{} encourages branch disentanglement of pretrained tokens into task-oriented content and subject-consistent style, then performs downstream prediction from the content branch. It instantiates this factorize-before-classify design with recursive Gaussian refinement, temporal-variance regularization, subject-aware contrastive learning, and reconstruction. Experiments across multiple EEG datasets and four EFM backbones show that \modelnickname{} improves matched fine-tuning baselines. Further ablations and analyses support the intended content/style branch organization.
\end{abstract}

\vspace*{-1.0em}
\suppressfloats[t]
\section{Introduction}
\label{intro}
\begin{figure}[t]
    \centering
    \includegraphics[width=\textwidth]{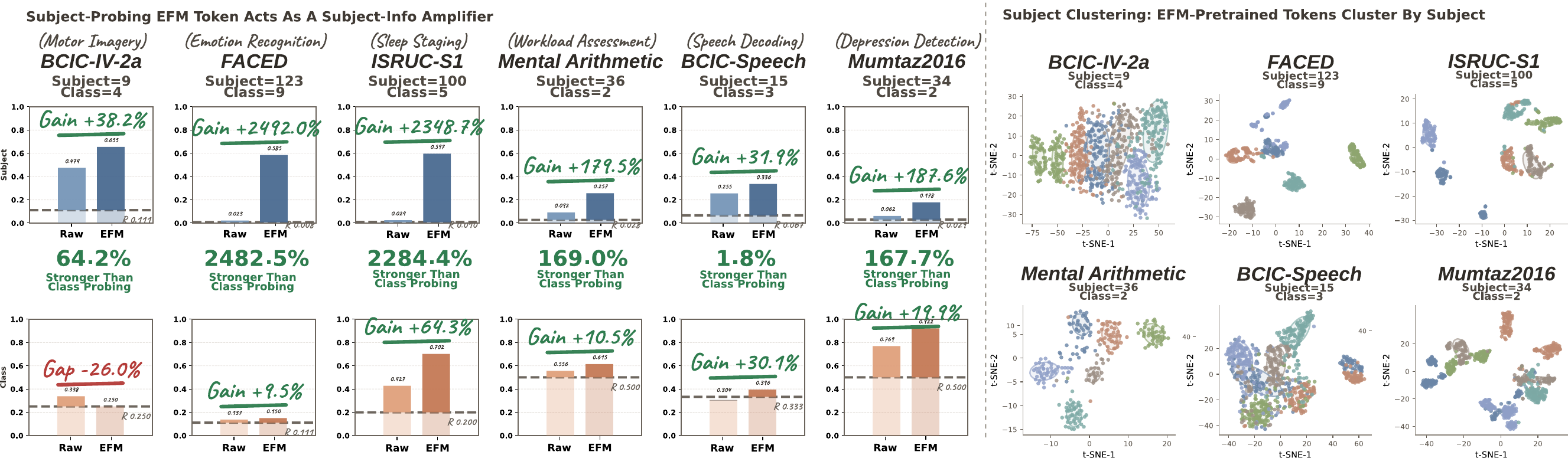}
    \caption{\textbf{Pretrained EFM embeddings amplify subject information.} Frozen CBraMod embeddings improve linear subject probing over raw EEG and form subject-dependent clusters. Task probes use subject-disjoint splits; subject probes use held-out trials from known subjects.}
    \label{fig:probe_results}
\end{figure}

Electroencephalography (EEG) provides high-temporal-resolution signals for a wide range of neural decoding tasks, such as emotion recognition \citep{wei-long_zheng_investigating_2015,chen_large_2023,liu_comprehensive_2024}, motor imagery \citep{tangermann_review_2012,schalk_eeg_2009,lee_eeg_2019}, and sleep staging \citep{jia_graphsleepnet_2020,zhou_bit-mamsleep_2024,zhou_introducing_2025}. Recent EEG foundation models (EFMs) improve scalability by pretraining on large multi-subject corpora and learning reusable representations that transfer across downstream settings \citep{zhou_csbrain_2025,jiang_large_2023,ma_codebrain_2025}. This paradigm has substantially improved the transferability and generalizability of EEG models \citep{jiang_large_2023,ma_codebrain_2025}, and is reshaping standard downstream EEG fine-tuning practice.

Probing reveals a consistent pattern across the evaluated settings: \textit{\textbf{pretrained EFM embeddings make subject identity more linearly accessible.}} As shown in Figure~\ref{fig:probe_results}, a probing classifier recovers subject identity more easily from pretrained EFM embeddings than from raw EEG features, while class-label probing improves only modestly after EFM encoding. EFM encoding can therefore make subject-related structure easier to separate in the pretrained token space. We use the compact factorization \mixedprose{} $\rightarrow$ (\contentprose{}, \styleprose{}) to summarize this mixed space, where \mixedprose{} denotes mixed pretrained tokens, \contentprose{} task-content tokens, and \styleprose{} subject-style tokens. Raw EEG already carries implicit subject traces, and EFM pretraining can sharpen these directions inside the mixed representation. EEG recordings are influenced by stable individual factors such as anatomy, functional organization, electrode-skin impedance, and recording habits \citep{wu_transfer_2022,saha_intra-_2020,lopez_stability_2023,brookshire_data_2024}. Current EFM embeddings consequently form mixed representations containing both task and subject structure.

\begin{figure}[!b]
    \centering
    \includegraphics[width=\textwidth]{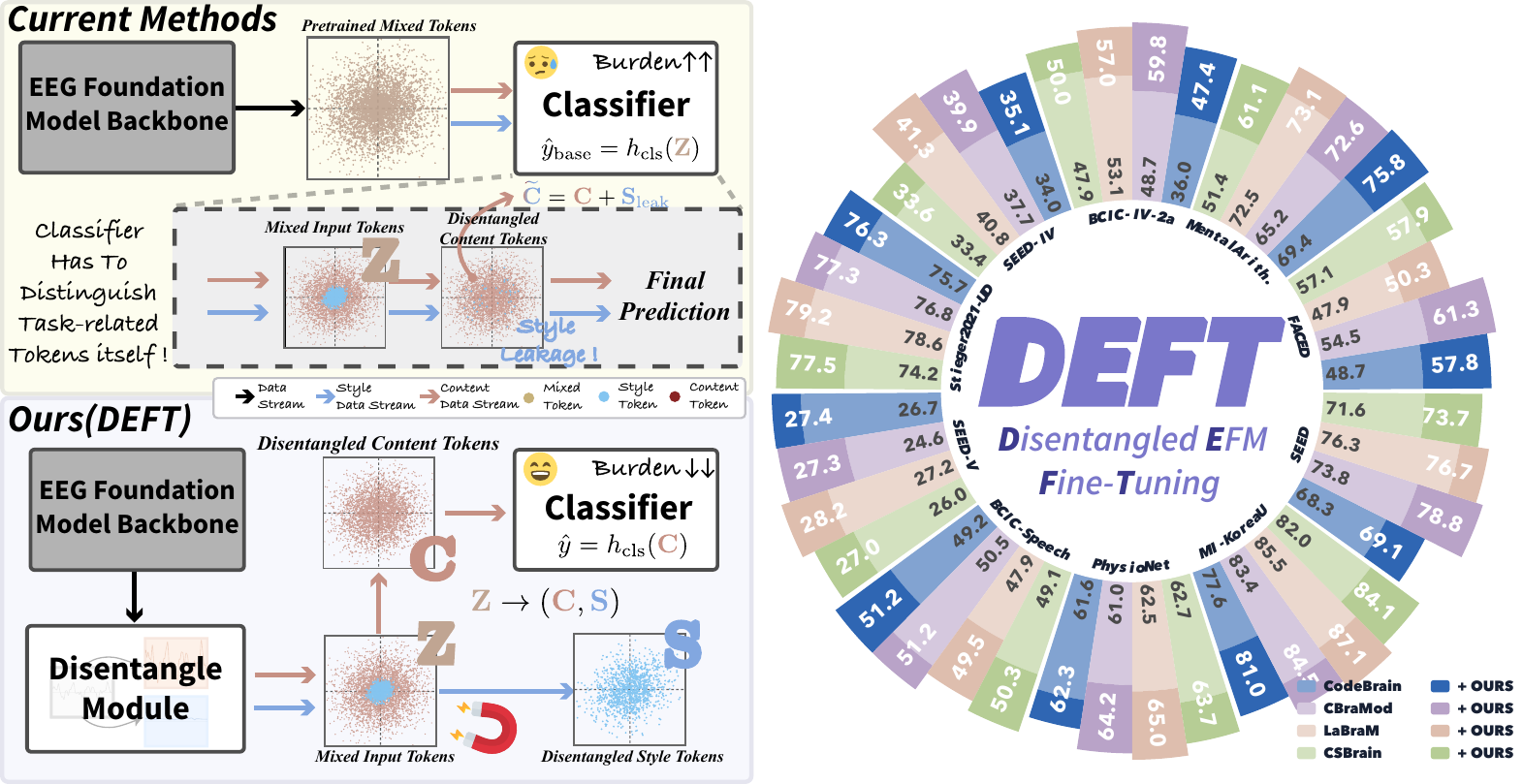}
    \caption{\textbf{Overview: motivation and empirical effect of \modelnickname{}.} \textit{Left}: standard EFM fine-tuning leaves the classifier to suppress subject-related variation inside mixed pretrained tokens, while \modelnickname{} models subject-consistent information before classification. \textit{Right}: an illustrative Balanced Accuracy snapshot across representative datasets and foundation models, showing that \modelnickname{} improves over vanilla fine-tuning in most displayed comparisons.}
    \label{img:intro}
\end{figure}

This amplification exposes a structural limitation of current EFM fine-tuning: \textit{\textbf{the strongest signal in a pretrained token may be the wrong signal for the downstream task.}} Figure~\ref{img:intro} (left) illustrates this burden. If pretraining sharpens subject structure more than class structure, the downstream classifier inherits a representation that is rich but misaligned. With only task labels as supervision, the prediction head must first discount easily accessible subject structure and then recover class-relevant content from the remaining mixed representation. This coupling is especially undesirable in \textit{\textbf{cross-subject settings}}, where subject cues are stable, easy to exploit, and poorly aligned with the desired decision rule. As a result, standard fine-tuning may spend part of its capacity correcting the pretrained representation before learning the classifier, which can obscure class-discriminative cues and weaken robustness in cross-subject EEG decoding \citep{wu_transfer_2022,shen_contrastive_2023}. This failure mode raises a critical question:

\smallskip
{\centering \textit{\textbf{How can we prevent amplified subject-related variation in pretrained EFM embeddings from burdening downstream classification?}}\par}
\smallskip

To address this problem, we propose \modelname{} (\modelnickname{}), a post-backbone disentanglement adapter for EFM fine-tuning. It operates on mixed pretrained tokens and forms a subject-style branch together with a task-content branch for downstream classification. The key idea is to absorb the stronger subject-related structure earlier in the representation, so that the final head no longer has to resolve the full mixture on its own. The two branches are refined by recursive Gaussian disentanglement, shaped by an intrinsic temporal-variance constraint, and regularized by task classification, subject-aware self-supervised contrast, and reconstruction. In this way, disentanglement is handled at the representation level and the classifier can focus more directly on class discrimination. This design gives the prediction head a cleaner working representation in practice while still preserving the subject branch as an explicit modeling target.

In our main evaluation, \modelnickname{} improves matched EFM fine-tuning baselines across multiple EEG task families and four foundation backbones. These results support the factorize-before-classify design: encouraging branch disentanglement into task-oriented content and subject-consistent style gives the prediction head a more useful representation for cross-subject EEG decoding. Additional ablations, representation analysis, and probing diagnostics suggest that representation-level adaptation contributes positively to downstream classification and provide analytic evidence consistent with our content/style interpretation.

\section{Preliminary}
\label{preliminary}
\subsection{EEG Foundation Model}
\label{sec:preliminary_efm}

Let $\mathbf{x} \in \mathbb{R}^{T \times N_{\mathrm{ch}}}$ denote an EEG trial with $T$ time steps and $N_{\mathrm{ch}}$ channels. A pretrained EEG foundation model (EFM) backbone $\mathcal{B}(\cdot)$ maps $\mathbf{x}$ to a token tensor
\begin{equation}
\mixedtokens = \mathcal{B}(\mathbf{x}) = \{z_{m,w}\}_{m=1,w=1}^{M,W} \in \mathbb{R}^{M \times W \times D},
\end{equation}
where $M$ is the latent channel-group dimension, $W$ is the number of temporal windows, and $D$ is the token width. We denote this mixed pretrained representation by \mixedprose{}. EFMs are pretrained on large multi-subject EEG corpora to learn transferable token representations for downstream adaptation.

\subsection{EFM Fine-Tuning}
\label{sec:preliminary_tokens}

Let $y \in \mathcal{Y}$ denote the task label and $u \in \mathcal{U}$ denote the subject identity. Standard fine-tuning attaches a downstream classification head $h_{\mathrm{cls}}(\cdot)$ to the pretrained tokens and predicts
\begin{equation}
\hat{y} = h_{\mathrm{cls}}(\mixedtokens), \qquad
\mathcal{L}_{\mathrm{cls}} = \mathrm{CE}(\hat{y}, y).
\end{equation}
In practice, $h_{\mathrm{cls}}(\cdot)$ is a lightweight prediction head that operates on pretrained tokens, typically through token aggregation followed by a linear or shallow MLP classifier.

In this work, \mixedprose{} is treated as a mixed representation space. Each token may encode both class-relevant neural dynamics and subject-related stable patterns, which forces standard fine-tuning to optimize $h_{\mathrm{cls}}(\cdot)$ on a representation where the two factors remain coupled. We therefore use \textbf{content} and \textbf{style} as functional branch roles in the pretrained token space. Content tokens \contentprose{} retain label-relevant brain-state structure and its temporal variation for subject-disjoint prediction. Style tokens \styleprose{} collect trial-stable residual structure organized primarily by subject identity. The prediction head operates on the content branch, while temporal-stability and subject-aware constraints shape the style branch.

\section{Methodology}
\label{method}
\label{sec:method}
This section presents \modelname{} (\modelnickname{}), a post-backbone disentanglement adapter for EEG foundation model (EFM) fine-tuning.

\subsection{Framework Overview}
\label{sec:framework_overview}

\textbf{\textit{Intuition:}} Pretrained EFM embeddings mix class-relevant dynamics with subject-specific bias. A direct classifier must learn the task decision while suppressing subject cues in the same token space. \modelnickname{} inserts an intermediate factorization step before prediction, so subject-consistent structure and task-discriminative variation can be shaped by separate training signals.

\begin{figure*}[t]
\centering
\includegraphics[width=\textwidth]{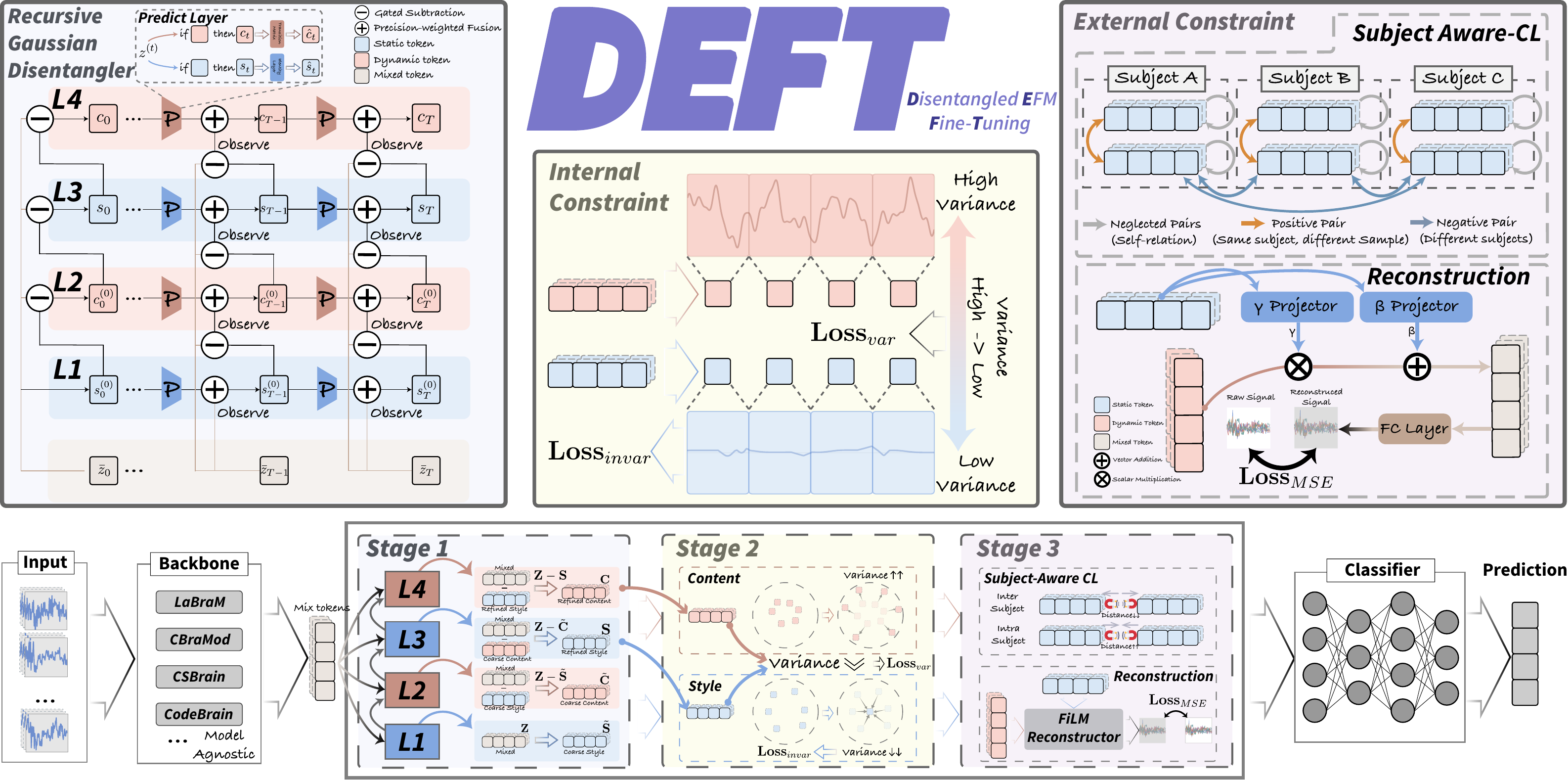}
\caption{Overall framework of \modelnickname{}. Mixed pretrained tokens are organized into task-content and subject-style branches, then optimized with temporal-variance regularization, classification, subject-aware contrastive learning, and reconstruction. The downstream classifier uses the content branch for prediction. Detailed processes of each stage are shown above the overall framework.}
\label{fig:method_overview}
\end{figure*}

In \modelnickname{}, content and style denote operational branch roles induced by the training objectives. The prediction head is applied only to \contentprose{}, so task gradients pass directly through the content branch. Subject-aware contrast, temporal stability, and reconstruction shape the style branch, while the content branch serves as the prediction interface. This factorize-before-classify design keeps cross-subject prediction focused on task-oriented content while modeling subject-linked nuisance variation explicitly.

\modelnickname{} is used as a post-backbone adapter. Starting from the mixed token tensor \mixedprose{}, we summarize it along the latent channel-group dimension and obtain a compact window-level sequence
\begin{equation}
\bar{z}_{w} = \frac{1}{M}\sum_{m=1}^{M} z_{m,w}, \qquad
\bar{\mixedtokens} = [\bar{z}_{1}, \bar{z}_{2}, \dots, \bar{z}_{W}] \in \mathbb{R}^{W \times D}.
\end{equation}
The framework then follows three stages. The Recursive Gaussian Disentangler first maps the compact mixed sequence into window-level branches, referred to as content and style according to their training roles:
\begin{equation}
(\mathbf{C}^{\mathrm{win}}, \mathbf{S}^{\mathrm{win}}) = \mathcal{G}_{\mathrm{dis}}(\bar{\mixedtokens}). 
\end{equation}
These window-level branches are then reinjected into the original token grid to form token-level content \contentprose{} and style \styleprose{}. The ordering loss $\mathcal{L}_{\mathrm{ord}}$ and style-stability loss $\mathcal{L}_{\mathrm{sty}}$ define the intrinsic constraint; the subject-aware contrastive loss $\mathcal{L}_{\mathrm{con}}$ and reconstruction loss $\mathcal{L}_{\mathrm{rec}}$ provide external anchors. Sections~\ref{sec:intrinsic_constraint}--\ref{sec:external_constraint} give their full definitions. The compact objective is
\begin{equation}
\begin{aligned}
\mathcal{L}_{\mathrm{int}} &=
\lambda_{\mathrm{ord}}\mathcal{L}_{\mathrm{ord}} +
\lambda_{\mathrm{sty}}\mathcal{L}_{\mathrm{sty}}, \quad
\mathcal{L}_{\mathrm{ext}} &=
\lambda_{\mathrm{con}}\mathcal{L}_{\mathrm{con}} +
\lambda_{\mathrm{rec}}\mathcal{L}_{\mathrm{rec}}, \quad
\mathcal{L} &= \mathcal{L}_{\mathrm{cls}} + \mathcal{L}_{\mathrm{int}} + \mathcal{L}_{\mathrm{ext}}.
\end{aligned}
\end{equation}
Together, these components form a task-content \textbf{Content Branch} and a subject-style \textbf{Style Branch}.

\subsection{Recursive Gaussian Disentangler}
\label{sec:gaussian_disentangler}

\textbf{\textit{Intuition:}} The Recursive Gaussian Disentangler performs iterative two-branch inference from counterpart-subtracted observations and temporal priors. It adapts the precision-weighted linear-Gaussian aggregation used in Xi-vector and the prior--observation fusion principle of Kalman filtering to pretrained EFM tokens \citep{lee_xivector_2021,kalman_new_1960}. Learned diagonal precision controls the contribution of token dimensions whose reliability varies with artifacts, electrode contact, and nonstationary brain activity. Alternating counterpart subtraction and asymmetric static/dynamic transitions then refine content and style estimates.

\noindent For each compact window $\bar{z}_w$, we derive observation statistics and remove the current counterpart estimate $r_w$:
\begin{equation}
e_w = f_{\mu}(\bar{z}_w), \quad \log \Lambda_w = f_{\lambda}(\bar{z}_w), \quad \tilde{e}_w = e_w - \gamma_w r_w,
\end{equation}
where $\gamma_w \in (0,1)$ is a learnable gate controlling counterpart removal.

\observationtitle{Observation 3.1}
Let $e_w$ be the observation for window $w$, and let $r_w$ be the current estimate of the counterpart branch. If the branch update uses the counterpart-subtracted observation $\tilde e_w=e_w-\gamma_w r_w$ with $\gamma_w\in(0,1)$, then its inner product with the counterpart estimate is reduced by $\gamma_w\|r_w\|_2^2$. Alternating this update for the two branches therefore makes each branch estimate depend on evidence after subtracting the current counterpart estimate. The full Gaussian update equations are given in Appendix~\ref{app:gaussian_details}.

\derivationstart
For a branch update with counterpart estimate $r_w$, direct substitution gives
$\langle \tilde e_w,r_w\rangle=\langle e_w,r_w\rangle-\gamma_w\|r_w\|_2^2$.
Thus, the observation used by the current branch has a smaller inner product with the counterpart estimate whenever $r_w\neq 0$ and $\gamma_w>0$. Alternating the updates gives $s_w\leftarrow \mathcal{G}_{\mathrm{sta}}(e_w-\gamma_w c_w)$ and $c_w\leftarrow \mathcal{G}_{\mathrm{dyn}}(e_w-\gamma_w s_w)$, so each branch is estimated after subtracting the current estimate of the other branch. The Gaussian fusion posterior then weights this counterpart-subtracted observation by precision, with $h_w=(\Pi_w+\tilde\Lambda_w)^{-1}(\Pi_wT_wh_{w-1}+\tilde\Lambda_w\tilde e_w)$ as detailed in Appendix~\ref{app:gaussian_details}. Recursive counterpart subtraction therefore biases each branch update toward lower overlap with its current counterpart estimate before downstream classification.

\notetitle{Note 3.1}
The key observation is alternating refinement: each branch is updated with access to its current counterpart estimate and with uncertainty-aware fusion.

The two branches share the same Gaussian update but use different transition operators. For the style branch, we use a static transition $T_w = I$. For the content branch, we use a data-dependent transition
\begin{equation}
T_w = G_w = \sum_{n=1}^{N} \alpha_{w,n} G_n, \qquad
\boldsymbol{\alpha}_w = \mathrm{Softmax}(g(h_{w-1})),
\end{equation}
where $\{G_n\}_{n=1}^{N}$ are learnable transition bases and $g(\cdot)$ is a lightweight gating network that selects their mixture. Starting from the mixed observation sequence, we perform four alternating inference steps:
\begin{equation}
\begin{aligned}
\left(s_w^{(0)}, P_w^{(0)}\right) &= \mathcal{G}_{\mathrm{sta}}(e_w,\Lambda_w), &
\left(c_w^{(0)}, Q_w^{(0)}\right) &= \mathcal{G}_{\mathrm{dyn}}(e_w,\Lambda_w; s_w^{(0)}, P_w^{(0)}), \\
\left(s_w, P_w\right) &= \mathcal{G}_{\mathrm{sta}}(e_w,\Lambda_w; c_w^{(0)}, Q_w^{(0)}), &
\left(c_w, Q_w\right) &= \mathcal{G}_{\mathrm{dyn}}(e_w,\Lambda_w; s_w, P_w).
\end{aligned}
\end{equation}
This yields window-level factors \textcolor{contentwine}{$\mathbf{C}^{\mathrm{win}}=[c_1,\dots,c_W]$} and \textcolor{styleblue}{$\mathbf{S}^{\mathrm{win}}=[s_1,\dots,s_W]$}, which are applied through residual reinjection into the original token grid:
\begin{equation}
z^{c}_{m,w} = (z_{m,w} - \bar{z}_w) + c_w, \qquad
z^{s}_{m,w} = (z_{m,w} - \bar{z}_w) + s_w.
\end{equation}
The residual term $z_{m,w}-\bar z_w$ preserves channel-group information when the shared window summary is replaced. It yields token-level content \textcolor{contentwine}{$\mathbf{C}=[z^{c}_{m,w}]_{m,w}$} and style \textcolor{styleblue}{$\mathbf{S}=[z^{s}_{m,w}]_{m,w}$}.

\subsection{Intrinsic Temporal-Variance Constraint}
\label{sec:intrinsic_constraint}

\textbf{\textit{Intuition:}} For the evaluated cross-subject tasks, we apply a task-dependent soft prior to learned window features: \textbf{content is encouraged to vary more across windows}, while \textbf{style is encouraged to remain stable} within a trial. This prior targets class-irrelevant subject bias while retaining stable evidence that predicts the task label in the content branch.

\observationtitle{Observation 3.2}
Let $\mathbf{H} = [h_1,\dots,h_W]$ be a window sequence and let
\begin{equation}
\bar{h} = \frac{1}{W}\sum_{w=1}^{W} h_w, \qquad
\mathrm{Var}(\mathbf{H}) =
\frac{1}{W}\sum_{w=1}^{W}\left\|h_w - \bar{h}\right\|_2^2.
\end{equation}
If \contentprose{} should preserve temporally varying task evidence and \styleprose{} should absorb trial-stable subject structure, then a variance-ordering loss with a style-stability penalty encourages the intended ordering under the margin constraint and directly reduces within-trial style variation.

\derivationstart
By definition, $\mathrm{Var}(\mathbf{H})$ measures within-trial deviation from the sequence mean. We implement temporal ordering and style stability with
\begin{equation}
\mathcal{L}_{\mathrm{ord}} =
\max\left(0, m + \mathrm{Var}(\styletokens) - \mathrm{Var}(\contenttokens)\right), \qquad
\mathcal{L}_{\mathrm{sty}} = \mathrm{Var}(\styletokens).
\end{equation}
The hinge term is zero exactly when $\mathrm{Var}(\contenttokens)\geq \mathrm{Var}(\styletokens)+m$. When it is active, $\partial \mathrm{Var}(\mathbf{H})/\partial h_w=\frac{2}{W}(h_w-\bar h)$, so gradient descent on $\mathcal{L}_{\mathrm{ord}}$ moves content windows away from their mean and style windows toward their mean. The second term gives the same contraction direction for style by directly penalizing $\mathrm{Var}(\styletokens)$. Together, the two losses encode the intended design principle: content is encouraged to retain stronger window-level variation, while style is encouraged to remain stable within a trial.

\notetitle{Note 3.2}
The intrinsic constraint is a label-free, task-dependent soft prior for settings where subject-linked structure acts as a nuisance factor for cross-subject prediction.

\subsection{External Constraints}
\label{sec:external_constraint}

\textbf{\textit{Intuition:}} The intrinsic variance prior shapes temporal behavior. Two external anchors assign functional roles to the branches: subject-aware contrast organizes style by subject identity, and reconstruction preserves information across the split.

During training, a subject-balanced sampler provides at least one same-subject positive for every anchor in the six main datasets, so the style contrastive loss uses both same-subject positives and cross-subject negatives. Anchors without same-subject positives are excluded from $\mathcal{L}_{\mathrm{con}}$ and retain their classification, temporal-variance, and reconstruction losses.

\observationtitle{Observation 3.3}
Let \contentprose{} denote the prediction branch, let \textcolor{styleblue}{$\mathbf{S}^{\mathrm{win}}=[s_1,\dots,s_W]$} denote the window-level style sequence, and let $\bar{z}_w$ denote the mixed window representation. Consider
\begin{equation}
\mathcal{I}_{+}=\{i:|P(i)|>0\}, \qquad
\mathcal{L}_{\mathrm{con}} =
\sum_{i\in \mathcal{I}_{+}}
\frac{-1}{|P(i)|}
\sum_{p \in P(i)}
\log
\frac{
\exp(\mathrm{sim}(\mathbf{s}^{\ast}_i,\mathbf{s}^{\ast}_p)/\tau)
}{
\sum_{a \neq i}
\exp(\mathrm{sim}(\mathbf{s}^{\ast}_i,\mathbf{s}^{\ast}_a)/\tau)
},
\end{equation}
and
\begin{equation}
\mathbf{s}^{\ast} = \frac{1}{W}\sum_{w=1}^{W} s_w, \qquad
\lambda, b = g_{\mathrm{film}}(\mathbf{s}^{\ast}), \qquad
\hat{z}_w = \lambda \odot c_w + b, \qquad
\mathcal{L}_{\mathrm{rec}} =
\frac{1}{W}\sum_{w=1}^{W}\left\|\hat{z}_w - \bar{z}_w\right\|_2^2.
\end{equation}
Here, $P(i)$ denotes same-subject positives for sample $i$, and anchors without positives are excluded by $\mathcal{I}_{+}$. Then $\mathcal{L}_{\mathrm{con}}$ organizes \styleprose{} by subject consistency, while $\mathcal{L}_{\mathrm{rec}}$ discourages information loss in the content-style split.

\derivationstart
For $i\in\mathcal{I}_{+}$, minimizing $\mathcal{L}_{\mathrm{con}}$ increases $\mathrm{sim}(\mathbf{s}^{\ast}_i,\mathbf{s}^{\ast}_p)$ for $p\in P(i)$ and decreases its relative value against $\mathrm{sim}(\mathbf{s}^{\ast}_i,\mathbf{s}^{\ast}_a)$ for $a\neq i$. Thus, same-subject style summaries are pulled together in \styleprose{}. For reconstruction, $\mathbf{s}^{\ast}$ is pooled across windows and only supplies sample-level FiLM parameters, while $c_w$ remains the window-varying input to $\hat z_w$. Minimizing $\mathcal{L}_{\mathrm{rec}}$ therefore requires the content branch to retain reconstructible window-level information that the style summary alone cannot provide. Thus, contrast anchors the style branch by subject, while reconstruction keeps the split informative.

\notetitle{Note 3.3}
In the overview, $\mathcal{L}_{\mathrm{ext}}$ contains only the contrastive and reconstruction terms. Thus, the external constraints assist representation learning before the final prediction head. Subject identities are used only during training for $\mathcal{L}_{\mathrm{con}}$ and are not required at inference time.
The full training objective is
\begin{equation}
\label{eq:total_loss}
\mathcal{L} =
\mathrm{CE}\left(h_{\mathrm{cls}}(\contenttokens), y\right)
+ \lambda_{\mathrm{ord}} \mathcal{L}_{\mathrm{ord}}
+ \lambda_{\mathrm{sty}} \mathcal{L}_{\mathrm{sty}}
+ \lambda_{\mathrm{con}} \mathcal{L}_{\mathrm{con}}
+ \lambda_{\mathrm{rec}} \mathcal{L}_{\mathrm{rec}},
\end{equation}
where $\lambda_{\mathrm{ord}}$, $\lambda_{\mathrm{sty}}$, $\lambda_{\mathrm{con}}$, and $\lambda_{\mathrm{rec}}$ balance the intrinsic and external constraints.

\section{Experiments}
\label{experiments}
\definecolor{gainorange}{RGB}{255,236,204}
\definecolor{lossblue}{RGB}{221,235,247}
\newcommand{\bestperf}[1]{\textbf{#1}}
\newcommand{\bestcell}[1]{\bestperf{#1}}
\newcommand{\posdelta}[1]{\cellcolor{gainorange}#1}
\newcommand{\negdelta}[1]{\cellcolor{lossblue}#1}
\newcommand{\deltacell}[1]{\posdelta{#1}}

\subsection{Experimental Setup}
\label{sec:exp_setup}

\textbf{Dataset and Backbone Setting:} We evaluate \modelnickname{} on four representative EEG foundation backbones: CodeBrain \citep{ma_codebrain_2025}, CBraMod \citep{wang_cbramod_2024}, LaBraM \citep{jiang_large_2023}, and CSBrain \citep{zhou_csbrain_2025}. The main text reports six representative downstream datasets under subject-disjoint evaluation: motor imagery (BCIC-IV-2a \citep{tangermann_review_2012}), emotion recognition (FACED \citep{chen_large_2023}), sleep staging (ISRUC-S1), workload assessment (MentalArith. \citep{zyma_electroencephalograms_2019}), speech decoding (BCIC-Speech), and MDD classification (Mumtaz2016). These datasets cover distinct EEG task families, and complete results across the full evaluation suite are provided in Appendix~\ref{app:complete_main_results}. Preprocessing details are provided in Appendix~\ref{app:dataset_details}.

\textbf{Model Setting:} We use the public released checkpoints of CodeBrain, CBraMod, LaBraM, and CSBrain and do not re-train the EFM pretraining stage. Following previous EEG foundation-model fine-tuning protocols, the pretrained EFM backbone, the downstream head, and any added adaptation modules are trainable during downstream training. The vanilla baseline optimizes the backbone and prediction head, while \modelnickname{} follows the same backbone update policy and jointly optimizes the disentanglement adapter. This keeps the comparison within the same checkpoint, split, and trainable encoder fine-tuning setting. The adapter adds only a small parameter overhead relative to the original fine-tuning model, including the pretrained backbone and downstream prediction head, as summarized in Appendix~\ref{app:parameter_overhead}. Detailed model parameters are described in Appendix~\ref{app:implementation_details}.

\textbf{Training and Environment Setting:} Experiments were conducted on NVIDIA RTX 4090 GPUs. Detailed training hyperparameters are provided in Appendix~\ref{app:training_hparams}.

\textbf{Evaluation Metrics:} Following prior EEG foundation model evaluations \citep{jiang_large_2023,wang_cbramod_2024,ma_codebrain_2025,zhou_csbrain_2025}, we report Balanced Accuracy (B-ACC) for all tasks, ROC-AUC/PR-AUC for binary tasks, and Kappa/weighted F1 for multi-class tasks. In the main table, all metrics report multi-seed mean and standard deviation over available seeds, and the complete records under seed 0 are provided in Appendix~\ref{app:complete_main_results}.

\begin{table}[!t]
\centering
\scriptsize
\renewcommand{\arraystretch}{1.05}
\caption{Main results on six representative downstream datasets. All metrics report multi-seed mean$\pm$std over available seeds. Bold marks the best value in each metric column, and Delta is the absolute improvement of \modelnickname{} over vanilla fine-tuning. Chance B-ACC is 0.250, 0.111, 0.200, 0.500, 0.333, and 0.500 for BCIC-IV-2a, FACED, ISRUC-S1, MentalArith., BCIC-Speech, and Mumtaz2016, respectively.}
\label{tab:main_result}
\setlength{\tabcolsep}{2.0pt}
\resizebox{0.997\textwidth}{!}{%
\begin{tabular}{@{}llccccccccc@{}}
\toprule
\textbf{Model} & \textbf{Setting} & \multicolumn{3}{c}{\footnotesize\textbf{BCIC-IV-2a}} & \multicolumn{3}{c}{\footnotesize\textbf{FACED}} & \multicolumn{3}{c}{\footnotesize\textbf{ISRUC-S1}} \\
\cmidrule(lr){3-5} \cmidrule(lr){6-8} \cmidrule(l){9-11}
& & B-ACC & Kappa & F1 & B-ACC & Kappa & F1 & B-ACC & Kappa & F1 \\
\midrule
\multirow{3}{*}{CodeBrain} & Vanilla & 0.3678$\pm$0.0220 & 0.1570$\pm$0.0293 & 0.2980$\pm$0.0369 & 0.5720$\pm$0.0047 & 0.5156$\pm$0.0055 & 0.5719$\pm$0.0048 & 0.7803$\pm$0.0071 & 0.7366$\pm$0.0084 & 0.7943$\pm$0.0059 \\
    & \modelnickname{} & 0.4621$\pm$0.0040 & 0.2828$\pm$0.0054 & 0.4318$\pm$0.0061 & 0.5870$\pm$0.0041 & 0.5297$\pm$0.0041 & 0.5799$\pm$0.0031 & 0.7829$\pm$0.0042 & 0.7391$\pm$0.0069 & 0.7981$\pm$0.0019 \\
& Delta & \deltacell{+0.0943} & \deltacell{+0.1258} & \deltacell{+0.1339} & \deltacell{+0.0150} & \deltacell{+0.0141} & \deltacell{+0.0080} & \deltacell{+0.0026} & \deltacell{+0.0025} & \deltacell{+0.0039} \\
\midrule
\multirow{3}{*}{CBraMod} & Vanilla & 0.4630$\pm$0.0327 & 0.2840$\pm$0.0437 & 0.4332$\pm$0.0519 & 0.5387$\pm$0.0108 & 0.4770$\pm$0.0118 & 0.5387$\pm$0.0100 & 0.7767$\pm$0.0094 & 0.7514$\pm$0.0109 & 0.8052$\pm$0.0086 \\
    & \modelnickname{} & 0.5579$\pm$0.0310 & 0.4105$\pm$0.0414 & 0.5522$\pm$0.0341 & \bestperf{0.6005$\pm$0.0101} & \bestperf{0.5467$\pm$0.0122} & \bestperf{0.6002$\pm$0.0130} & 0.7818$\pm$0.0080 & 0.7285$\pm$0.0108 & 0.7906$\pm$0.0110 \\
& Delta & \deltacell{+0.0949} & \deltacell{+0.1265} & \deltacell{+0.1191} & \deltacell{+0.0618} & \deltacell{+0.0697} & \deltacell{+0.0615} & \deltacell{+0.0052} & \negdelta{-0.0229} & \negdelta{-0.0145} \\
\midrule
\multirow{3}{*}{LaBraM} & Vanilla & 0.5434$\pm$0.0050 & 0.3912$\pm$0.0066 & 0.5349$\pm$0.0031 & 0.5336$\pm$0.0058 & 0.4727$\pm$0.0059 & 0.5351$\pm$0.0047 & 0.7730$\pm$0.0197 & 0.7350$\pm$0.0216 & 0.7773$\pm$0.0315 \\
    & \modelnickname{} & \bestperf{0.5671$\pm$0.0032} & \bestperf{0.4228$\pm$0.0043} & \bestperf{0.5601$\pm$0.0064} & 0.5409$\pm$0.0086 & 0.4820$\pm$0.0101 & 0.5462$\pm$0.0096 & \bestperf{0.8051$\pm$0.0086} & \bestperf{0.7792$\pm$0.0072} & \bestperf{0.8288$\pm$0.0149} \\
& Delta & \deltacell{+0.0237} & \deltacell{+0.0316} & \deltacell{+0.0252} & \deltacell{+0.0072} & \deltacell{+0.0093} & \deltacell{+0.0112} & \deltacell{+0.0321} & \deltacell{+0.0442} & \deltacell{+0.0515} \\
\midrule
\multirow{3}{*}{CSBrain} & Vanilla & 0.5148$\pm$0.0233 & 0.3530$\pm$0.0311 & 0.4980$\pm$0.0256 & 0.4169$\pm$0.2162 & 0.3441$\pm$0.2433 & 0.4031$\pm$0.2598 & 0.7738$\pm$0.0048 & 0.7297$\pm$0.0121 & 0.7819$\pm$0.0064 \\
    & \modelnickname{} & 0.5448$\pm$0.0226 & 0.3931$\pm$0.0301 & 0.5345$\pm$0.0289 & 0.5748$\pm$0.0033 & 0.5195$\pm$0.0037 & 0.5820$\pm$0.0055 & 0.7871$\pm$0.0052 & 0.7471$\pm$0.0078 & 0.8031$\pm$0.0057 \\
& Delta & \deltacell{+0.0301} & \deltacell{+0.0401} & \deltacell{+0.0365} & \deltacell{+0.1579} & \deltacell{+0.1754} & \deltacell{+0.1789} & \deltacell{+0.0133} & \deltacell{+0.0175} & \deltacell{+0.0212} \\
\bottomrule
\end{tabular}}
\vspace{2pt}
\resizebox{0.997\textwidth}{!}{%
\begin{tabular}{@{}llccccccccc@{}}
\toprule
\textbf{Model} & \textbf{Setting} & \multicolumn{3}{c}{\footnotesize\textbf{MentalArith.}} & \multicolumn{3}{c}{\footnotesize\textbf{BCIC-Speech}} & \multicolumn{3}{c}{\footnotesize\textbf{Mumtaz2016}} \\
\cmidrule(lr){3-5} \cmidrule(lr){6-8} \cmidrule(l){9-11}
& & B-ACC & ROC-AUC & PR-AUC & B-ACC & Kappa & F1 & B-ACC & ROC-AUC & PR-AUC \\
\midrule
\multirow{3}{*}{CodeBrain} & Vanilla & 0.6351$\pm$0.0635 & 0.7594$\pm$0.0618 & 0.5663$\pm$0.0433 & 0.5716$\pm$0.0304 & 0.3574$\pm$0.0457 & 0.5659$\pm$0.0416 & 0.9045$\pm$0.0045 & 0.9775$\pm$0.0021 & 0.9814$\pm$0.0014 \\
    & \modelnickname{} & 0.7267$\pm$0.0319 & 0.8080$\pm$0.0223 & 0.5785$\pm$0.0356 & 0.6188$\pm$0.0188 & 0.4281$\pm$0.0282 & 0.6158$\pm$0.0226 & 0.9090$\pm$0.0093 & 0.9701$\pm$0.0123 & 0.9770$\pm$0.0080 \\
& Delta & \deltacell{+0.0917} & \deltacell{+0.0486} & \deltacell{+0.0122} & \deltacell{+0.0472} & \deltacell{+0.0707} & \deltacell{+0.0499} & \deltacell{+0.0044} & \negdelta{-0.0074} & \negdelta{-0.0045} \\
\midrule
\multirow{3}{*}{CBraMod} & Vanilla & 0.6316$\pm$0.0148 & 0.7224$\pm$0.0161 & 0.5378$\pm$0.0166 & 0.6662$\pm$0.0251 & 0.4993$\pm$0.0376 & 0.6661$\pm$0.0254 & 0.8531$\pm$0.0439 & 0.9764$\pm$0.0096 & 0.9779$\pm$0.0097 \\
    & \modelnickname{} & \bestperf{0.7292$\pm$0.0059} & \bestperf{0.8142$\pm$0.0190} & \bestperf{0.6131$\pm$0.0503} & \bestperf{0.6894$\pm$0.0172} & \bestperf{0.5341$\pm$0.0258} & \bestperf{0.6903$\pm$0.0187} & \bestperf{0.9473$\pm$0.0142} & \bestperf{0.9919$\pm$0.0018} & \bestperf{0.9935$\pm$0.0018} \\
& Delta & \deltacell{+0.0976} & \deltacell{+0.0918} & \deltacell{+0.0754} & \deltacell{+0.0232} & \deltacell{+0.0348} & \deltacell{+0.0242} & \deltacell{+0.0942} & \deltacell{+0.0155} & \deltacell{+0.0156} \\
\midrule
\multirow{3}{*}{LaBraM} & Vanilla & 0.7198$\pm$0.0507 & 0.7857$\pm$0.0574 & 0.6088$\pm$0.0932 & 0.5630$\pm$0.0321 & 0.3444$\pm$0.0481 & 0.5551$\pm$0.0414 & 0.9109$\pm$0.0096 & 0.9808$\pm$0.0054 & 0.9836$\pm$0.0042 \\
    & \modelnickname{} & 0.7236$\pm$0.0149 & 0.7753$\pm$0.0262 & 0.5228$\pm$0.0478 & 0.6069$\pm$0.0096 & 0.4104$\pm$0.0145 & 0.5980$\pm$0.0096 & 0.9266$\pm$0.0025 & 0.9849$\pm$0.0021 & 0.9862$\pm$0.0017 \\
& Delta & \deltacell{+0.0038} & \negdelta{-0.0104} & \negdelta{-0.0860} & \deltacell{+0.0440} & \deltacell{+0.0659} & \deltacell{+0.0428} & \deltacell{+0.0157} & \deltacell{+0.0041} & \deltacell{+0.0026} \\
\midrule
\multirow{3}{*}{CSBrain} & Vanilla & 0.5510$\pm$0.0374 & 0.7266$\pm$0.0067 & 0.4860$\pm$0.0425 & 0.5958$\pm$0.0103 & 0.3937$\pm$0.0154 & 0.5913$\pm$0.0103 & 0.8939$\pm$0.0089 & 0.9804$\pm$0.0044 & 0.9835$\pm$0.0037 \\
    & \modelnickname{} & 0.6319$\pm$0.0157 & 0.7252$\pm$0.0363 & 0.4201$\pm$0.0479 & 0.6299$\pm$0.0109 & 0.4448$\pm$0.0163 & 0.6300$\pm$0.0131 & 0.9220$\pm$0.0180 & 0.9869$\pm$0.0059 & 0.9884$\pm$0.0047 \\
& Delta & \deltacell{+0.0809} & \negdelta{-0.0013} & \negdelta{-0.0659} & \deltacell{+0.0341} & \deltacell{+0.0511} & \deltacell{+0.0387} & \deltacell{+0.0281} & \deltacell{+0.0066} & \deltacell{+0.0049} \\
\bottomrule
\end{tabular}}
\vspace{-0.6em}
\end{table}

\subsection{Main Results}
\label{sec:main_result}

\textbf{Scope of the main comparison.} Table~\ref{tab:main_result} is the primary quantitative comparison: six representative subject-disjoint datasets, four EFM backbones, and multi-seed results over available seeds. Appendix~\ref{app:complete_main_results} provides complete seed-0 records for completeness. Across the 24 matched backbone--dataset cells, \modelnickname{} improves mean B-ACC by 0.0460, with a two-way backbone/dataset clustered-bootstrap 95\% confidence interval of $[0.0240, 0.0745]$. The gains appear across CodeBrain, CBraMod, LaBraM, and CSBrain.

\textbf{(a) Improvements span diverse task families.} The largest mean B-ACC gains occur on MentalArith., BCIC-IV-2a, and FACED. Together with the gains on sleep staging, speech decoding, and MDD classification, these results cover distinct EEG prediction settings.

\textbf{(b) Complementary metrics support the same overall trend.} Most Kappa, F1, ROC-AUC, and PR-AUC values improve, and Table~\ref{tab:main_result} retains every observed decrease. The confidence intervals for the reviewer-identified secondary-metric differences are reported in Appendix~\ref{app:secondary_metric_uncertainty}.

\textbf{(c) Backbone-level gains vary in magnitude.} Averaged over the six representative datasets, B-ACC gains are +0.0628 for CBraMod, +0.0574 for CSBrain, +0.0425 for CodeBrain, and +0.0211 for LaBraM. Appendix~\ref{app:rgd_controls} reports additional controls on CBraMod, LaBraM, and REVE.

\subsection{Ablation Study}
\label{sec:ablation}

We examine each component with fixed seeds 0--2 on CBraMod. Table~\ref{tab:ablation} copies the multi-seed Full values from Table~\ref{tab:main_result} and reports the corresponding multi-seed B-ACC ablations. The component effects are task dependent: the largest reductions appear on BCIC-IV-2a, MentalArith., BCIC-Speech, and Mumtaz2016, while the changes on FACED and ISRUC-S1 are smaller. Appendix~\ref{app:complete_ablation} retains the complete seed-0 metric records, and Appendix~\ref{app:rgd_controls} provides cross-backbone and objective-combination controls.

\textbf{(a) Component sensitivity depends on the task.} BCIC-IV-2a is especially sensitive to the invariance, variance-ratio, and Gaussian components; MentalArith. and BCIC-Speech show substantial reductions across all ablations.

\textbf{(b) The objectives provide complementary branch supervision.} The objective-combination control in Appendix~\ref{app:objective_combinations} shows task-dependent interactions among the variance, contrastive, and reconstruction terms.

\begin{table}[!t]
\centering
\scriptsize
\renewcommand{\arraystretch}{1.02}
\caption{Multi-seed CBraMod B-ACC ablations on the six representative datasets. Ablations use fixed seeds 0--2; Full is copied from Table~\ref{tab:main_result}.}
\label{tab:ablation}
\setlength{\tabcolsep}{2.2pt}
\resizebox{0.997\textwidth}{!}{%
\begin{tabular}{@{}lccccccc@{}}
\toprule
\textbf{Dataset} & \textbf{Chance} & \textbf{Full} & \textbf{w/o StyleCL} & \textbf{w/o Invar.} & \textbf{w/o VarRatio} & \textbf{w/o Gaussian} & \textbf{w/o Rec.} \\
\midrule
BCIC-IV-2a & 0.250 & 0.5579$\pm$0.0310 & 0.4638$\pm$0.0170 & 0.3973$\pm$0.0664 & 0.3808$\pm$0.0071 & 0.3782$\pm$0.0198 & 0.4615$\pm$0.0477 \\
FACED & 0.111 & 0.6005$\pm$0.0101 & 0.5931$\pm$0.0072 & 0.5992$\pm$0.0011 & 0.5976$\pm$0.0027 & 0.5998$\pm$0.0087 & 0.6042$\pm$0.0066 \\
ISRUC-S1 & 0.200 & 0.7818$\pm$0.0080 & 0.7897$\pm$0.0012 & 0.7849$\pm$0.0019 & 0.7838$\pm$0.0121 & 0.7856$\pm$0.0057 & 0.7782$\pm$0.0090 \\
MentalArith. & 0.500 & 0.7292$\pm$0.0059 & 0.5830$\pm$0.0367 & 0.6549$\pm$0.0478 & 0.5687$\pm$0.0527 & 0.5958$\pm$0.0321 & 0.6135$\pm$0.0295 \\
BCIC-Speech & 0.333 & 0.6894$\pm$0.0172 & 0.4908$\pm$0.0034 & 0.4858$\pm$0.0082 & 0.4870$\pm$0.0109 & 0.4972$\pm$0.0076 & 0.4776$\pm$0.0120 \\
Mumtaz2016 & 0.500 & 0.9473$\pm$0.0142 & 0.8286$\pm$0.0678 & 0.9049$\pm$0.0475 & 0.7581$\pm$0.1772 & 0.9126$\pm$0.0114 & 0.8760$\pm$0.0329 \\
\bottomrule
\end{tabular}}
\end{table}

For BCIC-IV-2a, removing invariance changes B-ACC by $-0.1606$, with a Welch 95\% confidence interval of $[-0.2999,-0.0213]$ across the three seeds.

\subsection{Representation Analysis}
\label{sec:analysis_visualization}

The main and ablation results show that \modelnickname{} improves downstream decoding, and we further examine whether the content branch also changes the geometry of the mixed representation in the intended direction. We use a post-hoc \textit{within-subject class separation} score after subject centering and PCA50. Higher values indicate that task-class centroids are farther apart relative to within-class dispersion after subject-level offsets have been controlled. The detailed computation protocol is provided in Appendix~\ref{app:representation_geometry_metric}.

\begin{figure}[t!]
\centering
\vspace{-0.65em}
\includegraphics[width=0.965\textwidth]{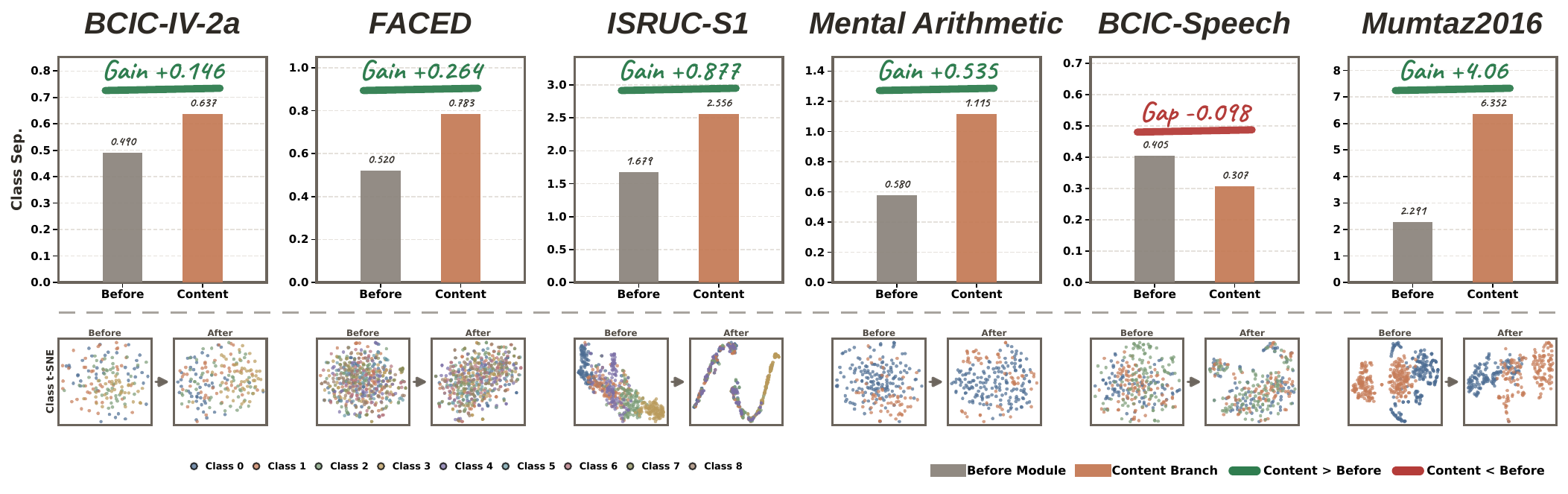}
\caption{\textbf{Representation geometry before and after disentanglement.} Top: subject-controlled within-subject class separation after subject centering, where higher values indicate clearer class structure within each subject. Bottom: class t-SNE visualizations of the same before-module and content-branch representations.}
\label{fig:analysis_visualization}
\vspace{-1.15em}
\end{figure}

The top of Figure~\ref{fig:analysis_visualization} shows that the content branch improves the class-separation score on five of the six representative datasets: BCIC-IV-2a (0.490 $\rightarrow$ 0.637, +0.146), FACED (0.520 $\rightarrow$ 0.783, +0.264), ISRUC-S1 (1.679 $\rightarrow$ 2.556, +0.877), MentalArith. (0.580 $\rightarrow$ 1.115, +0.535), and Mumtaz2016 (2.291 $\rightarrow$ 6.352, +4.061). These increases indicate that, after subject-level offsets are removed, the content branch usually makes class centroids farther apart relative to within-class dispersion.

BCIC-Speech is the only case where this post-hoc geometry score is slightly reduced (0.405 $\rightarrow$ 0.307), despite metric gains for all four backbones in Table~\ref{tab:main_result}. This mismatch suggests that speech-decoding cues may be localized across channels and windows, making them less visible as global PCA clusters. Overall, the content branch usually produces clearer task-content geometry, but the geometry gain can vary with task characteristics.

Direct frozen linear probes provide a complementary measurement in Appendix~\ref{app:disentanglement_comparison}: content yields higher task-probe B-ACC and style yields higher subject-probe B-ACC in all six evaluated settings. This quantifies the functional specialization of the two branches.

\section{Related Work}
\label{related_work}
\label{sec:related}

\textbf{EEG Foundation Models.} Encoder-based EFMs such as BIOT, EEGPT, LaBraM, CBraMod, CodeBrain, and CSBrain pretrain reusable EEG encoders with masked reconstruction, neural token prediction, or structured spatiotemporal modeling objectives \citep{yang_biot_2023,wang_eegpt_2024,jiang_large_2023,wang_cbramod_2024,ma_codebrain_2025,zhou_csbrain_2025}. Decoder-based or language-connected EFMs provide a complementary direction: NeuroLM connects EEG tokens to language-model instruction tuning, and ECHO casts EEG modeling as sequence-to-sequence prediction with support examples for contextual adaptation \citep{jiang_neurolm_2025,liu_echo_2025}. For encoder-based EFMs used in downstream classification, the common adaptation recipe remains vanilla fine-tuning: attach a task head and update the pretrained backbone under task supervision. \modelnickname{} instead studies a tailored fine-tuning adapter for this encoder-based setting, factorizing mixed pretrained tokens before prediction while remaining applicable to EFMs that expose token embeddings.

\textbf{Disentanglement Learning.} Disentangled representation learning separates factors such as content and style, stable and dynamic components, or identity and task information, with applications in speech and time-series representations \citep{hsu_unsupervised_2017,li_disentangled_2018,liu_disentangling_2023,oublal_disentangling_2024}. In EEG, patient-independent seizure detection has used adversarial or patient-invariant objectives to reduce patient-specific interference \citep{ganin_domain-adversarial_2016,zhang_adversarial_2020,zhang_cross-patient_2024}. Existing approaches learn subject-invariant representations through adversarial, alignment, or sequential factorization objectives. \modelnickname{} applies factorization after a pretrained EFM backbone and assigns task prediction and subject consistency to separate token branches.

\section{Conclusion}
\label{Conclusion}
We introduce \modelnickname{}, a disentanglement-enhanced fine-tuning framework for EEG foundation models that targets a practical failure mode of EFM adaptation: pretrained embeddings can make subject-related structure highly accessible to the downstream classifier, even when that structure is not the intended target of cross-subject decoding. \modelnickname{} inserts a post-backbone factorize-before-classify adapter that decomposes the mixed representation into prediction-oriented content and subject-consistent style, then drives downstream prediction from the content branch. Experiments across multiple EEG task families and four EFM backbones show that \modelnickname{} improves matched fine-tuning baselines, while ablation, representation analysis, and content/style probing support the intended branch organization. Appendix checks cover one additional EFM, one independent MDD dataset, and one MOMENT/UCI HAR setting. Overall, these results support disentangling pretrained EFM embeddings for content-driven cross-subject EEG decoding.


\newpage
\bibliographystyle{plain}
\bibliography{neurips_2026}

@article{tangermann_review_2012,
	title = {Review of the {BCI} {Competition} {IV}},
	volume = {6},
	issn = {1662-4548},
	url = {http://journal.frontiersin.org/article/10.3389/fnins.2012.00055/abstract},
	doi = {10.3389/fnins.2012.00055},
	urldate = {2026-04-26},
	journal = {Frontiers in Neuroscience},
	author = {Tangermann, Michael and Müller, Klaus-Robert and Aertsen, Ad and Birbaumer, Niels and Braun, Christoph and Brunner, Clemens and Leeb, Robert and Mehring, Carsten and Miller, Kai J. and Müller-Putz, Gernot R. and Nolte, Guido and Pfurtscheller, Gert and Preissl, Hubert and Schalk, Gerwin and Schlögl, Alois and Vidaurre, Carmen and Waldert, Stephan and Blankertz, Benjamin},
	year = {2012},
}

@article{chen_large_2023,
	title = {A {Large} {Finer}-grained {Affective} {Computing} {EEG} {Dataset}},
	volume = {10},
	issn = {2052-4463},
	url = {https://www.nature.com/articles/s41597-023-02650-w},
	doi = {10.1038/s41597-023-02650-w},
	language = {en},
	number = {1},
	urldate = {2026-04-26},
	journal = {Scientific Data},
	author = {Chen, Jingjing and Wang, Xiaobin and Huang, Chen and Hu, Xin and Shen, Xinke and Zhang, Dan},
	month = oct,
	year = {2023},
	pages = {740},
}

@misc{schalk_eeg_2009,
	title = {{EEG} {Motor} {Movement}/{Imagery} {Dataset}},
	url = {https://physionet.org/content/eegmmidb/},
	doi = {10.13026/C28G6P},
	urldate = {2026-04-26},
	publisher = {physionet.org},
	author = {Schalk, Gerwin and McFarland, Dennis J and Hinterberger, Thilo and Birbaumer, Niels and Wolpaw, Jonathan R},
	year = {2009},
}

@article{wei-long_zheng_investigating_2015,
	title = {Investigating {Critical} {Frequency} {Bands} and {Channels} for {EEG}-{Based} {Emotion} {Recognition} with {Deep} {Neural} {Networks}},
	volume = {7},
	copyright = {https://ieeexplore.ieee.org/Xplorehelp/downloads/license-information/IEEE.html},
	issn = {1943-0604, 1943-0612},
	url = {http://ieeexplore.ieee.org/document/7104132/},
	doi = {10.1109/TAMD.2015.2431497},
	number = {3},
	urldate = {2026-04-26},
	journal = {IEEE Transactions on Autonomous Mental Development},
	author = {{Wei-Long Zheng} and {Bao-Liang Lu}},
	month = sep,
	year = {2015},
	pages = {162--175},
}

@article{lee_eeg_2019,
	title = {{EEG} dataset and {OpenBMI} toolbox for three {BCI} paradigms: an investigation into {BCI} illiteracy},
	volume = {8},
	copyright = {http://creativecommons.org/licenses/by/4.0/},
	issn = {2047-217X},
	shorttitle = {{EEG} dataset and {OpenBMI} toolbox for three {BCI} paradigms},
	url = {https://academic.oup.com/gigascience/article/doi/10.1093/gigascience/giz002/5304369},
	doi = {10.1093/gigascience/giz002},
	language = {en},
	number = {5},
	urldate = {2026-04-26},
	journal = {GigaScience},
	author = {Lee, Min-Ho and Kwon, O-Yeon and Kim, Yong-Jeong and Kim, Hong-Kyung and Lee, Young-Eun and Williamson, John and Fazli, Siamac and Lee, Seong-Whan},
	month = may,
	year = {2019},
	pages = {giz002},
}

@article{zyma_electroencephalograms_2019,
	title = {Electroencephalograms during {Mental} {Arithmetic} {Task} {Performance}},
	volume = {4},
	issn = {2306-5729},
	url = {https://www.mdpi.com/2306-5729/4/1/14},
	doi = {10.3390/data4010014},
	language = {en},
	number = {1},
	urldate = {2026-04-26},
	journal = {Data},
	author = {Zyma, Igor and Tukaev, Sergii and Seleznov, Ivan and Kiyono, Ken and Popov, Anton and Chernykh, Mariia and Shpenkov, Oleksii},
	month = jan,
	year = {2019},
	pages = {14},
}

@misc{ganin_domain-adversarial_2016,
	title = {Domain-{Adversarial} {Training} of {Neural} {Networks}},
	url = {http://arxiv.org/abs/1505.07818},
	doi = {10.48550/arXiv.1505.07818},
	language = {en},
	urldate = {2026-04-26},
	publisher = {arXiv},
	author = {Ganin, Yaroslav and Ustinova, Evgeniya and Ajakan, Hana and Germain, Pascal and Larochelle, Hugo and Laviolette, François and Marchand, Mario and Lempitsky, Victor},
	month = may,
	year = {2016},
	note = {arXiv:1505.07818 [stat]},
}

@misc{chen_infogan_2016,
	title = {{InfoGAN}: {Interpretable} {Representation} {Learning} by {Information} {Maximizing} {Generative} {Adversarial} {Nets}},
	shorttitle = {{InfoGAN}},
	url = {http://arxiv.org/abs/1606.03657},
	doi = {10.48550/arXiv.1606.03657},
	language = {en},
	urldate = {2026-04-26},
	publisher = {arXiv},
	author = {Chen, Xi and Duan, Yan and Houthooft, Rein and Schulman, John and Sutskever, Ilya and Abbeel, Pieter},
	month = jun,
	year = {2016},
	note = {arXiv:1606.03657 [cs]},
}

@misc{kumar_variational_2018,
	title = {Variational {Inference} of {Disentangled} {Latent} {Concepts} from {Unlabeled} {Observations}},
	url = {http://arxiv.org/abs/1711.00848},
	doi = {10.48550/arXiv.1711.00848},
	language = {en},
	urldate = {2026-04-26},
	publisher = {arXiv},
	author = {Kumar, Abhishek and Sattigeri, Prasanna and Balakrishnan, Avinash},
	month = dec,
	year = {2018},
	note = {arXiv:1711.00848 [cs]},
}

@misc{chen_isolating_2019,
	title = {Isolating {Sources} of {Disentanglement} in {Variational} {Autoencoders}},
	url = {http://arxiv.org/abs/1802.04942},
	doi = {10.48550/arXiv.1802.04942},
	language = {en},
	urldate = {2026-04-26},
	publisher = {arXiv},
	author = {Chen, Ricky T. Q. and Li, Xuechen and Grosse, Roger and Duvenaud, David},
	month = apr,
	year = {2019},
	note = {arXiv:1802.04942 [cs]},
}

@misc{kim_disentangling_2019,
	title = {Disentangling by {Factorising}},
	url = {http://arxiv.org/abs/1802.05983},
	doi = {10.48550/arXiv.1802.05983},
	language = {en},
	urldate = {2026-04-26},
	publisher = {arXiv},
	author = {Kim, Hyunjik and Mnih, Andriy},
	month = jul,
	year = {2019},
	note = {arXiv:1802.05983 [stat]},
}

@article{shen_contrastive_2023,
	title = {Contrastive {Learning} of {Subject}-{Invariant} {EEG} {Representations} for {Cross}-{Subject} {Emotion} {Recognition}},
	volume = {14},
	copyright = {https://ieeexplore.ieee.org/Xplorehelp/downloads/license-information/IEEE.html},
	issn = {1949-3045, 2371-9850},
	url = {https://ieeexplore.ieee.org/document/9748967/},
	doi = {10.1109/TAFFC.2022.3164516},
	language = {en},
	number = {3},
	urldate = {2026-04-26},
	journal = {IEEE Transactions on Affective Computing},
	author = {Shen, Xinke and Liu, Xianggen and Hu, Xin and Zhang, Dan and Song, Sen},
	month = jul,
	year = {2023},
	pages = {2496--2511},
}

@misc{yang_biot_2023,
	title = {{BIOT}: {Cross}-data {Biosignal} {Learning} in the {Wild}},
	shorttitle = {{BIOT}},
	url = {http://arxiv.org/abs/2305.10351},
	doi = {10.48550/arXiv.2305.10351},
	language = {en},
	urldate = {2026-04-26},
	publisher = {arXiv},
	author = {Yang, Chaoqi and Westover, M. Brandon and Sun, Jimeng},
	month = may,
	year = {2023},
	note = {arXiv:2305.10351 [eess]},
}

@misc{liu_comprehensive_2024,
	title = {A {Comprehensive} {Survey} on {EEG}-{Based} {Emotion} {Recognition}: {A} {Graph}-{Based} {Perspective}},
	shorttitle = {A {Comprehensive} {Survey} on {EEG}-{Based} {Emotion} {Recognition}},
	url = {http://arxiv.org/abs/2408.06027},
	doi = {10.48550/arXiv.2408.06027},
	language = {en},
	urldate = {2026-04-26},
	publisher = {arXiv},
	author = {Liu, Chenyu and Zhou, Xinliang and Wu, Yihao and Ding, Yi and Zhai, Liming and Wang, Kun and Jia, Ziyu and Liu, Yang},
	month = aug,
	year = {2024},
	note = {arXiv:2408.06027 [eess]},
}

@misc{zhou_brain_2025,
	title = {Brain {Foundation} {Models}: {A} {Survey} on {Advancements} in {Neural} {Signal} {Processing} and {Brain} {Discovery}},
	copyright = {arXiv.org perpetual, non-exclusive license},
	shorttitle = {Brain {Foundation} {Models}},
	url = {https://arxiv.org/abs/2503.00580},
	doi = {10.48550/ARXIV.2503.00580},
	language = {en},
	urldate = {2026-04-26},
	publisher = {arXiv},
	author = {Zhou, Xinliang and Liu, Chenyu and Chen, Zhisheng and Wang, Kun and Ding, Yi and Jia, Ziyu and Wen, Qingsong},
	year = {2025},
}

@misc{locatello_challenging_2019,
	title = {Challenging {Common} {Assumptions} in the {Unsupervised} {Learning} of {Disentangled} {Representations}},
	url = {http://arxiv.org/abs/1811.12359},
	doi = {10.48550/arXiv.1811.12359},
	language = {en},
	urldate = {2026-04-27},
	publisher = {arXiv},
	author = {Locatello, Francesco and Bauer, Stefan and Lucic, Mario and Rätsch, Gunnar and Gelly, Sylvain and Schölkopf, Bernhard and Bachem, Olivier},
	month = jun,
	year = {2019},
	note = {arXiv:1811.12359 [cs]},
}

@misc{zhou_bit-mamsleep_2024,
	title = {{BiT}-{MamSleep}: {Bidirectional} {Temporal} {Mamba} for {EEG} {Sleep} {Staging}},
	shorttitle = {{BiT}-{MamSleep}},
	url = {http://arxiv.org/abs/2411.01589},
	doi = {10.48550/arXiv.2411.01589},
	language = {en},
	urldate = {2026-04-27},
	publisher = {arXiv},
	author = {Zhou, Xinliang and Han, Yuzhe and Chen, Zhisheng and Liu, Chenyu and Ding, Yi and Jia, Ziyu and Liu, Yang},
	month = nov,
	year = {2024},
	note = {arXiv:2411.01589 [eess]},
}

@inproceedings{jia_graphsleepnet_2020,
	address = {Yokohama, Japan},
	title = {{GraphSleepNet}: {Adaptive} {Spatial}-{Temporal} {Graph} {Convolutional} {Networks} for {Sleep} {Stage} {Classification}},
	isbn = {978-0-9992411-6-5},
	shorttitle = {{GraphSleepNet}},
	url = {https://www.ijcai.org/proceedings/2020/184},
	doi = {10.24963/ijcai.2020/184},
	language = {en},
	urldate = {2026-04-28},
	booktitle = {Proceedings of the {Twenty}-{Ninth} {International} {Joint} {Conference} on {Artificial} {Intelligence}},
	publisher = {International Joint Conferences on Artificial Intelligence Organization},
	author = {Jia, Ziyu and Lin, Youfang and Wang, Jing and Zhou, Ronghao and Ning, Xiaojun and He, Yuanlai and Zhao, Yaoshuai},
	month = jul,
	year = {2020},
	pages = {1324--1330},
}

@misc{zhou_introducing_2025,
	title = {Introducing {Multimodal} {Paradigm} for {Learning} {Sleep} {Staging} {PSG} via {General}-{Purpose} {Model}},
	url = {http://arxiv.org/abs/2509.22810},
	doi = {10.48550/arXiv.2509.22810},
	language = {en},
	urldate = {2026-04-28},
	publisher = {arXiv},
	author = {Zhou, Jianheng and Liu, Chenyu and Zhou, Jinan and Ding, Yi and Liu, Yang and Luo, Haoran and Jia, Ziyu and Zhou, Xinliang},
	month = sep,
	year = {2025},
	note = {arXiv:2509.22810 [eess]},
}

@article{oublal_disentangling_2024,
	title = {Disentangling {Time} {Series} {Representations} via {Contrastive} {Independence}-of-{Support} on l-{Variational} {Inference}},
	language = {en},
	author = {Oublal, Khalid and Ladjal, Saıd and Benhaiem, David and Le-borgne, Emmanuel and Roueff, Francois},
	year = {2024},
}

@misc{liu_disentangling_2023,
	title = {Disentangling {Voice} and {Content} with {Self}-{Supervision} for {Speaker} {Recognition}},
	url = {http://arxiv.org/abs/2310.01128},
	doi = {10.48550/arXiv.2310.01128},
	language = {en},
	urldate = {2026-05-01},
	publisher = {arXiv},
	author = {Liu, Tianchi and Lee, Kong Aik and Wang, Qiongqiong and Li, Haizhou},
	month = nov,
	year = {2023},
	note = {arXiv:2310.01128 [eess]},
}

@misc{liu_echo_2025,
	title = {{ECHO}: {Toward} {Contextual} {Seq2Seq} {Paradigms} in {Large} {EEG} {Models}},
	shorttitle = {{ECHO}},
	url = {http://arxiv.org/abs/2509.22556},
	doi = {10.48550/arXiv.2509.22556},
	language = {en},
	urldate = {2026-05-01},
	publisher = {arXiv},
	author = {Liu, Chenyu and Deng, Yuqiu and Liu, Tianyu and Zhou, Jinan and Zhou, Xinliang and Jia, Ziyu and Ding, Yi},
	month = sep,
	year = {2025},
	note = {arXiv:2509.22556 [cs]},
}

@inproceedings{jiang_neurolm_2025,
	title = {{NeuroLM}: {A} {Universal} {Multi}-{task} {Foundation} {Model} for {Bridging} the {Gap} between {Language} and {EEG} {Signals}},
	booktitle = {International Conference on Learning Representations},
	url = {https://openreview.net/forum?id=Io9yFt7XH7},
	author = {Jiang, Weibang and Wang, Yansen and Lu, Bao-Liang and Li, Dongsheng},
	year = {2025},
}

@misc{zhang_adversarial_2020,
	title = {Adversarial {Representation} {Learning} for {Robust} {Patient}-{Independent} {Epileptic} {Seizure} {Detection}},
	url = {http://arxiv.org/abs/1909.10868},
	doi = {10.48550/arXiv.1909.10868},
	language = {en},
	urldate = {2026-05-01},
	publisher = {arXiv},
	author = {Zhang, Xiang and Yao, Lina and Dong, Manqing and Liu, Zhe and Zhang, Yu and Li, Yong},
	month = feb,
	year = {2020},
	note = {arXiv:1909.10868 [eess]},
}

@article{zhang_cross-patient_2024,
	title = {Cross-patient automatic epileptic seizure detection using patient-adversarial neural networks with spatio-temporal {EEG} augmentation},
	volume = {89},
	issn = {1746-8094},
	url = {https://www.sciencedirect.com/science/article/pii/S1746809423010972},
	doi = {10.1016/j.bspc.2023.105664},
	urldate = {2026-05-01},
	journal = {Biomedical Signal Processing and Control},
	author = {Zhang, Zongpeng and Ji, Taoyun and Xiao, Mingqing and Wang, Wen and Yu, Guojing and Lin, Tong and Jiang, Yuwu and Zhou, Xiaohua and Lin, Zhouchen},
	month = mar,
	year = {2024},
	pages = {105664},
}

@article{saha_intra-_2020,
	title = {Intra- and {Inter}-subject {Variability} in {EEG}-{Based} {Sensorimotor} {Brain} {Computer} {Interface}: {A} {Review}},
	volume = {13},
	issn = {1662-5188},
	shorttitle = {Intra- and {Inter}-subject {Variability} in {EEG}-{Based} {Sensorimotor} {Brain} {Computer} {Interface}},
	url = {https://www.frontiersin.org/article/10.3389/fncom.2019.00087/full},
	doi = {10.3389/fncom.2019.00087},
	urldate = {2026-05-01},
	journal = {Frontiers in Computational Neuroscience},
	author = {Saha, Simanto and Baumert, Mathias},
	month = jan,
	year = {2020},
	pages = {87},
}

@article{lopez_stability_2023,
	title = {Stability, change, and reliable individual differences in electroencephalography measures: {A} lifespan perspective on progress and opportunities},
	volume = {275},
	issn = {10538119},
	shorttitle = {Stability, change, and reliable individual differences in electroencephalography measures},
	url = {https://linkinghub.elsevier.com/retrieve/pii/S1053811923002628},
	doi = {10.1016/j.neuroimage.2023.120116},
	language = {en},
	urldate = {2026-05-01},
	journal = {NeuroImage},
	author = {Lopez, K.L. and Monachino, A.D. and Vincent, K.M. and Peck, F.C. and Gabard-Durnam, L.J.},
	month = jul,
	year = {2023},
	pages = {120116},
}

@article{brookshire_data_2024,
	title = {Data leakage in deep learning studies of translational {EEG}},
	volume = {18},
	issn = {1662-453X},
	url = {https://www.frontiersin.org/articles/10.3389/fnins.2024.1373515/full},
	doi = {10.3389/fnins.2024.1373515},
	urldate = {2026-05-01},
	journal = {Frontiers in Neuroscience},
	author = {Brookshire, Geoffrey and Kasper, Jake and Blauch, Nicholas M. and Wu, Yunan Charles and Glatt, Ryan and Merrill, David A. and Gerrol, Spencer and Yoder, Keith J. and Quirk, Colin and Lucero, Ché},
	month = may,
	year = {2024},
	pages = {1373515},
}

@article{wang_eegpt_2024,
	title = {{EEGPT}: {Pretrained} {Transformer} for {Universal} and {Reliable} {Representation} of {EEG} {Signals}},
	volume = {37},
	shorttitle = {{EEGPT}},
	url = {https://proceedings.neurips.cc/paper_files/paper/2024/hash/4540d267eeec4e5dbd9dae9448f0b739-Abstract-Conference.html},
	doi = {10.52202/079017-1239},
	language = {en},
	urldate = {2026-05-06},
	journal = {Advances in Neural Information Processing Systems},
	author = {Wang, Guagnyu and Liu, Wenchao and He, Yuhong and Xu, Cong and Ma, Lin and Li, Haifeng},
	month = dec,
	year = {2024},
	pages = {39249--39280},
}

@inproceedings{jiang_large_2023,
	title = {Large {Brain} {Model} for {Learning} {Generic} {Representations} with {Tremendous} {EEG} {Data} in {BCI}},
	booktitle = {International Conference on Learning Representations},
	url = {https://openreview.net/forum?id=QzTpTRVtrP},
	language = {en},
	urldate = {2026-05-06},
	author = {Jiang, Weibang and Zhao, Liming and Lu, Bao-liang},
	month = oct,
	year = {2023},
}

@inproceedings{ma_codebrain_2025,
	title = {{CodeBrain}: {Bridging} {Decoupled} {Tokenizer} and {Multi}-{Scale} {Architecture} for {EEG} {Foundation} {Model}},
	shorttitle = {{CodeBrain}},
	booktitle = {International Conference on Learning Representations},
	url = {https://openreview.net/forum?id=msJgEkjwh5},
	language = {en},
	urldate = {2026-05-06},
	author = {Ma, Jingying and Wu, Feng and Lin, Qika and Xing, Yucheng and Liu, Chenyu and Jia, Ziyu and Feng, Mengling},
	month = oct,
	year = {2025},
}

@inproceedings{wang_cbramod_2024,
	title = {{CBraMod}: {A} {Criss}-{Cross} {Brain} {Foundation} {Model} for {EEG} {Decoding}},
	shorttitle = {{CBraMod}},
	booktitle = {International Conference on Learning Representations},
	url = {https://openreview.net/forum?id=NPNUHgHF2w},
	language = {en},
	urldate = {2026-05-06},
	author = {Wang, Jiquan and Zhao, Sha and Luo, Zhiling and Zhou, Yangxuan and Jiang, Haiteng and Li, Shijian and Li, Tao and Pan, Gang},
	month = oct,
	year = {2024},
}

@inproceedings{zhou_csbrain_2025,
	title = {{CSBrain}: {A} {Cross}-scale {Spatiotemporal} {Brain} {Foundation} {Model} for {EEG} {Decoding}},
	shorttitle = {{CSBrain}},
	booktitle = {Advances in Neural Information Processing Systems},
	url = {https://openreview.net/forum?id=agcXjEHmyW},
	language = {en},
	urldate = {2026-05-06},
	author = {Zhou, Yuchen and Wu, Jiamin and Ren, Zichen and Yao, Zhouheng and Lu, Weiheng and Peng, Kunyu and Zheng, Qihao and Song, Chunfeng and Ouyang, Wanli and Gou, Chao},
	month = oct,
	year = {2025},
}

@article{wu_transfer_2022,
	title = {Transfer {Learning} for {EEG}-{Based} {Brain}–{Computer} {Interfaces}: {A} {Review} of {Progress} {Made} {Since} 2016},
	volume = {14},
	copyright = {https://ieeexplore.ieee.org/Xplorehelp/downloads/license-information/IEEE.html},
	issn = {2379-8920, 2379-8939},
	shorttitle = {Transfer {Learning} for {EEG}-{Based} {Brain}–{Computer} {Interfaces}},
	url = {https://ieeexplore.ieee.org/document/9134411/},
	doi = {10.1109/TCDS.2020.3007453},
	number = {1},
	urldate = {2026-05-06},
	journal = {IEEE Transactions on Cognitive and Developmental Systems},
	author = {Wu, Dongrui and Xu, Yifan and Lu, Bao-Liang},
	month = mar,
	year = {2022},
	pages = {4--19},
}

@article{lawhern2018eegnet,
  title={EEGNet: a compact convolutional neural network for EEG-based brain--computer interfaces},
  author={Lawhern, Vernon J and Solon, Amelia J and Waytowich, Nicholas R and Gordon, Stephen M and Hung, Chou P and Lance, Brent J},
  journal={Journal of neural engineering},
  volume={15},
  number={5},
  pages={056013},
  year={2018}
}

@inproceedings{liu2024vbh,
  title={VBH-GNN: Variational Bayesian Heterogeneous Graph Neural Networks for Cross-subject Emotion Recognition.},
  author={Liu, Chenyu and Zhou, Xinliang and Zhu, Zhengri and Zhai, Liming and Jia, Ziyu and Liu, Yang},
  booktitle={The Twelfth International Conference on Learning Representations},
  year={2024}
}

@inproceedings{elreve,
  title={REVE: A Foundation Model for EEG-Adapting to Any Setup with Large-Scale Pretraining on 25,000 Subjects},
  author={El Ouahidi, Yassine and Lys, Jonathan and Th{\"o}lke, Philipp and Farrugia, Nicolas and Pasdeloup, Bastien and Gripon, Vincent and Jerbi, Karim and Lioi, Giulia},
  booktitle={The Thirty-ninth Annual Conference on Neural Information Processing Systems},
  year={2025}
}

@article{wang2026deeperbrain,
  title={DeeperBrain: A Neuro-Grounded EEG Foundation Model Towards Universal BCI},
  author={Wang, Jiquan and Zhao, Sha and Zhou, Yangxuan and Kang, Yiming and Li, Shijian and Pan, Gang},
  journal={arXiv preprint arXiv:2601.06134},
  year={2026}
}

@article{lu2025unimind,
  title={Unimind: Unleashing the power of llms for unified multi-task brain decoding},
  author={Lu, Weiheng and Song, Chunfeng and Wu, Jiamin and Zhu, Pengyu and Zhou, Yuchen and Mai, Weijian and Zheng, Qihao and Ouyang, Wanli},
  journal={arXiv preprint arXiv:2506.18962},
  year={2025}
}

@article{jiang2025elastiq,
  title={ELASTIQ: EEG-Language Alignment with Semantic Task Instruction and Querying},
  author={Jiang, Muyun and Zhang, Shuailei and Yang, Zhenjie and Wu, Mengjun and Jiang, Weibang and Guo, Zhiwei and Zhang, Wei and Liu, Rui and Zhang, Shangen and Li, Yong and others},
  journal={arXiv preprint arXiv:2509.24302},
  year={2025}
}

@article{kostas2021bendr,
  title={BENDR: Using transformers and a contrastive self-supervised learning task to learn from massive amounts of EEG data},
  author={Kostas, Demetres and Aroca-Ouellette, Stephane and Rudzicz, Frank},
  journal={Frontiers in Human Neuroscience},
  volume={15},
  pages={653659},
  year={2021},
  publisher={Frontiers Media SA}
}

@inproceedings{mohammadi2024eeg2rep,
  title={Eeg2rep: enhancing self-supervised eeg representation through informative masked inputs},
  author={Mohammadi Foumani, Navid and Mackellar, Geoffrey and Ghane, Soheila and Irtza, Saad and Nguyen, Nam and Salehi, Mahsa},
  booktitle={Proceedings of the 30th ACM SIGKDD Conference on Knowledge Discovery and Data Mining},
  pages={5544--5555},
  year={2024}
}

@article{wang2025eegmamba,
  title={Eegmamba: An eeg foundation model with mamba},
  author={Wang, Jiquan and Zhao, Sha and Luo, Zhiling and Zhou, Yangxuan and Li, Shijian and Pan, Gang},
  journal={Neural Networks},
  pages={107816},
  year={2025},
  publisher={Elsevier}
}

@inproceedings{doner2025luna,
  title={LUNA: Efficient and topology-agnostic foundation model for EEG signal analysis},
  author={D{\"o}ner, Berkay and Ingolfsson, Thorir Mar and Benini, Luca and Li, Yawei},
  booktitle={Advances in Neural Information Processing Systems},
  year={2025}
}

@article{ding2025brainpro,
  title={BrainPro: Towards Large-scale Brain State-aware EEG Representation Learning},
  author={Ding, Yi and Jiang, Muyun and Jiang, Weibang and Zhang, Shuailei and Zhou, Xinliang and Liu, Chenyu and Li, Shanglin and Li, Yong and Guan, Cuntai},
  journal={arXiv preprint arXiv:2509.22050},
  year={2025}
}

@inproceedings{pradeepkumar2025tokenizing,
  title={Tokenizing Single-Channel EEG with Time-Frequency Motif Learning},
  author={Pradeepkumar, Jathurshan and Piao, Xihao and Chen, Zheng and Sun, Jimeng},
  booktitle={The Fourteenth International Conference on Learning Representations},
  year={2026}
}

@inproceedings{chen2026uni,
  title={Uni-NTFM: A Unified Foundation Model for EEG Signal Representation Learning},
  author={Chen, Zhisheng and Zhang, Yingwei and Lan, Qizhen and Liu, Tianyu and Wang, Huacan and Ding, Yi and Jia, Ziyu and Chen, Ronghao and Wang, Kun and Zhou, Xinliang},
  booktitle={The Fourteenth International Conference on Learning Representations},
  year={2026}
}

@article{yang2025foundation,
  title={Foundation and Large-Scale AI Models in Neuroscience: A Comprehensive Review},
  author={Yang, Shihao and Huang, Xiying and Bernardo, Danilo and Ding, Jun-En and Michael, Andrew and Yang, Jingmei and Kwan, Patrick and Raj, Ashish and Liu, Feng},
  journal={arXiv preprint arXiv:2510.16658},
  year={2025}
}

@article{liu2026eeg,
  title={EEG Foundation Models: Progresses, Benchmarking, and Open Problems},
  author={Liu, Dingkun and Chen, Yuheng and Chen, Zhu and Cui, Zhenyao and Wen, Yaozhi and An, Jiayu and Luo, Jingwei and Wu, Dongrui},
  journal={arXiv preprint arXiv:2601.17883},
  year={2026}
}

@article{song2022eeg,
  title={EEG conformer: Convolutional transformer for EEG decoding and visualization},
  author={Song, Yonghao and Zheng, Qingqing and Liu, Bingchuan and Gao, Xiaorong},
  journal={IEEE Transactions on Neural Systems and Rehabilitation Engineering},
  volume={31},
  pages={710--719},
  year={2022},
  publisher={IEEE}
}

@article{ding2024eeg,
  title={EEG-Deformer: A dense convolutional transformer for brain-computer interfaces},
  author={Ding, Yi and Li, Yong and Sun, Hao and Liu, Rui and Tong, Chengxuan and Liu, Chenyu and Zhou, Xinliang and Guan, Cuntai},
  journal={IEEE Journal of Biomedical and Health Informatics},
  volume={29},
  number={3},
  pages={1909--1918},
  year={2024},
  publisher={IEEE}
}

@inproceedings{liu2024vsgt,
  title={VSGT: Variational Spatial and Gaussian Temporal Graph Models for EEG-based Emotion Recognition.},
  author={Liu, Chenyu and Zhou, Xinliang and Xiao, Jiaping and Zhu, Zhengri and Zhai, Liming and Jia, Ziyu and Liu, Yang},
  booktitle={IJCAI},
  pages={3078--3086},
  year={2024}
}

@inproceedings{chien2022maeeg,
  title={MAEEG: Masked Auto-encoder for EEG Representation Learning},
  author={Chien, Hsiang-Yun Sherry and Goh, Hanlin and Sandino, Christopher Michael and Cheng, Joseph Yitan},
  booktitle={NeurIPS 2022 Workshop on Learning from Time Series for Health},
  year={2022}
}

@article{gui2024eegmamba,
  title={EEGMamba: Bidirectional state space model with mixture of experts for EEG multi-task classification},
  author={Gui, Yiyu and Chen, MingZhi and Su, Yuqi and Luo, Guibo and Yang, Yuchao},
  journal={arXiv preprint arXiv:2407.20254},
  year={2024}
}

@inproceedings{fangneuript,
  title={NeurIPT: Foundation Model for Neural Interfaces},
  author={Fang, Zitao and LI, CHENXUAN and Hongting, Zhou and Yu, Shuyang and DU, Guodong and Qasem, Ashwaq and Lu, Yang and Li, Jing and Zhang, Junsong and Goh, Sim Kuan},
  booktitle={The Thirty-ninth Annual Conference on Neural Information Processing Systems},
  year={2025}
}

@article{cui2026brainrvq,
  title={BrainRVQ: A High-Fidelity EEG Foundation Model via Dual-Domain Residual Quantization and Hierarchical Autoregression},
  author={Cui, Mingzhe and Chen, Tao and Jiao, Yang and Wang, Yiqin and Xie, Lei and Pan, Yi and Mainardi, Luca},
  journal={arXiv preprint arXiv:2602.16951},
  year={2026}
}

@inproceedings{gui2024vector,
  title={Vector Quantization Pretraining for EEG Time Series with Random Projection and Phase Alignment.},
  author={Gui, Haokun and Li, Xiucheng and Chen, Xinyang},
  booktitle={ICML},
  year={2024}
}

@article{dongyeop2025diver,
  title={DIVER-0: A Fully Channel Equivariant EEG Foundation Model},
  author={Dongyeop Han, Danny and Lee, Ahhyun Lucy and Lee, Taeyang and Gwon, Yonghyeon and Lee, Sebin and Lee, Seongjin and Keetae Park, David and Yoo, Shinjae and Cha, Jiook and Chung, Chun Kee},
  journal={arXiv e-prints},
  pages={arXiv--2507},
  year={2025}
}

@article{zhou2025introducing,
  title={Introducing Multimodal Paradigm for Learning Sleep Staging PSG via General-Purpose Model},
  author={Zhou, Jianheng and Liu, Chenyu and Zhou, Jinan and Ding, Yi and Liu, Yang and Luo, Haoran and Jia, Ziyu and Zhou, Xinliang},
  journal={arXiv preprint arXiv:2509.22810},
  year={2025}
}

@article{Jia2025SeraSC,
  title={Sera: Separated Coarse-to-fine Representation Alignment for Cross-subject EEG-based Emotion Recognition},
  author={Zhihao Jia and Meiyan Xu and Jingyuan Wang and Ziyu Jia and Yong Li and Xin-qiu Zhou and Chenyu Liu and Junfeng Yao and Yi Ding},
  journal={Proceedings of the 33rd ACM International Conference on Multimedia},
  year={2025},
}

@article{lee_xivector_2021,
  title={{Xi-Vector} Embedding for Speaker Recognition},
  author={Lee, Kong Aik and Wang, Qiongqiong and Koshinaka, Takafumi},
  journal={IEEE Signal Processing Letters},
  volume={28},
  pages={1385--1389},
  year={2021}
}

@article{kalman_new_1960,
  title={A New Approach to Linear Filtering and Prediction Problems},
  author={Kalman, Rudolf E.},
  journal={Journal of Basic Engineering},
  volume={82},
  number={1},
  pages={35--45},
  year={1960}
}

@inproceedings{hsu_unsupervised_2017,
  title={Unsupervised Learning of Disentangled and Interpretable Representations from Sequential Data},
  author={Hsu, Wei-Ning and Zhang, Yu and Glass, James},
  booktitle={Advances in Neural Information Processing Systems},
  year={2017}
}

@inproceedings{li_disentangled_2018,
  title={Disentangled Sequential Autoencoder},
  author={Li, Yingzhen and Mandt, Stephan},
  booktitle={International Conference on Machine Learning},
  pages={5670--5679},
  year={2018}
}

@inproceedings{goswami_moment_2024,
  title={{MOMENT}: A Family of Open Time-Series Foundation Models},
  author={Goswami, Mononito and Szafer, Konrad and Choudhry, Arjun and Cai, Yifu and Li, Shuo and Dubrawski, Artur},
  booktitle={International Conference on Machine Learning},
  pages={16115--16152},
  year={2024}
}

@inproceedings{anguita_public_2013,
  title={A Public Domain Dataset for Human Activity Recognition Using Smartphones},
  author={Anguita, Davide and Ghio, Alessandro and Oneto, Luca and Parra, Xavier and Reyes-Ortiz, Jorge L.},
  booktitle={European Symposium on Artificial Neural Networks, Computational Intelligence and Machine Learning},
  year={2013}
}

@misc{cavanagh_eeg_2021,
  title={{EEG}: Depression Rest},
  author={Cavanagh, James F.},
  year={2021},
  howpublished={OpenNeuro dataset ds003478, version 1.1.0},
  doi={10.18112/openneuro.ds003478.v1.1.0}
}


\newpage
\appendix

\section{Detailed Related Work}
\label{app:related}
\label{app:related_work}

\subsection{Scalp EEG Foundation Models}

EEG modeling has moved from task-specific supervised decoders toward reusable pretrained backbones. Earlier deep pipelines, including compact CNNs, graph-based sleep staging, convolutional Transformers, dense Transformer variants, and variational spatiotemporal graph models, encode useful task-level inductive biases \citep{lawhern2018eegnet,jia_graphsleepnet_2020,song2022eeg,ding2024eeg,liu2024vsgt}. Self-supervised EEG representation learning further reduced annotation dependence through contrastive prediction, masked reconstruction, and informative masked modeling \citep{kostas2021bendr,chien2022maeeg,mohammadi2024eeg2rep}, setting the stage for larger-scale EEG pretraining.

Recent EEG foundation models (EFMs) scale this paradigm across heterogeneous EEG corpora and task families \citep{zhou_brain_2025,yang2025foundation,liu2026eeg}. Encoder-based EFMs such as BIOT, EEGPT, LaBraM, CBraMod, CodeBrain, and CSBrain learn transferable token representations with cross-data biosignal modeling, masked signal reconstruction, neural token prediction, criss-cross spatial-temporal attention, decoupled tokenization, or cross-scale spatiotemporal modeling \citep{yang_biot_2023,wang_eegpt_2024,jiang_large_2023,wang_cbramod_2024,ma_codebrain_2025,zhou_csbrain_2025}. Recent variants broaden this space with state-space backbones and mixture-of-experts routing \citep{wang2025eegmamba,gui2024eegmamba,chen2026uni,fangneuript}, large-scale or topology-agnostic pretraining \citep{elreve,doner2025luna,dongyeop2025diver}, neuroscience-inspired priors \citep{ding2025brainpro,wang2026deeperbrain}, and richer token interfaces such as vector quantization, time-frequency motifs, and residual quantization \citep{gui2024vector,pradeepkumar2025tokenizing,cui2026brainrvq}.

A related branch connects EEG representations with language or multimodal systems, including EEG-language foundation modeling, contextual sequence-to-sequence EEG modeling, instruction-style brain decoding, EEG-language alignment, and general-purpose sleep-staging interfaces \citep{jiang_neurolm_2025,liu_echo_2025,lu2025unimind,jiang2025elastiq,zhou2025introducing}. These studies expand EFM capacity and interfaces, while downstream classification still commonly feeds pretrained EFM embeddings directly to a task head and optimizes in the mixed representation space. \modelnickname{} targets this adaptation stage by inserting a representation-level adapter between the EFM and prediction head, factorizing the mixed representation into content and style before classification.

\subsection{Disentanglement and Subject-Related EEG Variation}

Disentangled representation learning organizes latent variables according to factors of variation, with representative methods based on information-theoretic generative modeling, variational objectives, total-correlation penalties, and factorial latent distributions \citep{chen_infogan_2016,kumar_variational_2018,chen_isolating_2019,kim_disentangling_2019}. Later analyses show that useful disentanglement depends on inductive biases, supervision, and downstream utility \citep{locatello_challenging_2019}. In sequential domains, FHVAE and DSAE separate sequence-level factors from segment-level dynamics, and later work isolates content/style factors in speech and time-series representations \citep{hsu_unsupervised_2017,li_disentangled_2018,liu_disentangling_2023,oublal_disentangling_2024}. \modelnickname{} extends factorized representation learning to downstream adaptation of pretrained EFM tokens.

Subject-related variation is a central source of difficulty in EEG. Stable individual differences, intra- and inter-subject variability, and leakage from subject-specific patterns can strongly affect apparent generalization \citep{saha_intra-_2020,lopez_stability_2023,brookshire_data_2024}. Existing EEG methods often seek invariant or aligned representations through transfer learning, domain-adversarial training, patient-adversarial objectives, subject-invariant contrastive learning, variational graph modeling, or coarse-to-fine cross-subject alignment \citep{wu_transfer_2022,ganin_domain-adversarial_2016,zhang_adversarial_2020,zhang_cross-patient_2024,shen_contrastive_2023,liu2024vbh,Jia2025SeraSC}.

\modelnickname{} operates on pretrained EFM embeddings and treats them as a mixed representation that may already amplify subject-related information. The content branch is optimized for downstream prediction and encouraged to preserve task-relevant temporal variation, while the style branch collects subject-consistent structure that remains stable within a trial. This yields a factorize-before-classify adaptation interface for EFMs.

\section{Limitations, Ethics, and Reproducibility}
\label{app:limitations_ethics_reproducibility}

\paragraph{Limitations.}
\modelnickname{} introduces additional loss weights and an inductive temporal-variance prior, which are selected under the downstream validation protocol. For trait-level or demographic targets, validation selects the temporal-prior strength so stable label-relevant evidence remains in the content branch. A dataset with one sample per training subject provides no same-subject contrastive pairs; the classification, variance, and reconstruction objectives remain applicable. Subject identities support training-time regularization and are omitted at inference time. Fully unsupervised disentanglement fine-tuning remains an important direction for future work.

\paragraph{Ethics and reproducibility.}
\modelnickname{} uses subject identities only during training to construct subject-aware contrastive pairs. Experiments use released EEG datasets and subject-disjoint downstream splits. The potential positive impact is more robust cross-subject EEG representation learning for research evaluation. Since EEG and MDD-related data may contain sensitive health-related information, the main risk is inappropriate clinical interpretation or privacy-sensitive use; \modelnickname{} should not be treated as a clinical diagnostic system without dataset-specific clinical validation. Code and configuration files are available at \url{https://github.com/wythedee/DEFT} to support reproducibility.

\paragraph{Compute resources.}
\label{app:compute_resources}
The reported experiments use independent single-GPU jobs on NVIDIA RTX 4090 GPUs with 24 GB memory. These jobs can be run in parallel on a multi-GPU machine; GPU-hours therefore measure aggregate compute, not required wall-clock time. Table~\ref{tab:app_compute_accounting} summarizes the compute needed to reproduce the main reported evaluation, the complete seed-0 evaluation suite, and the diagnostic experiments reported in the appendix.

\begin{table}[!t]
\centering
\small
\renewcommand{\arraystretch}{1.05}
\caption{Compute accounting for reproducing the reported experiments. All values are approximate.}
\label{tab:app_compute_accounting}
\setlength{\tabcolsep}{3.5pt}
\resizebox{\textwidth}{!}{%
\begin{tabular}{@{}lccc@{}}
\toprule
\textbf{Scope} & \textbf{Training fits} & \textbf{GPU / fit} & \textbf{GPU-hours} \\
\midrule
Main Table~\ref{tab:main_result} multi-seed reproduction & 120--150 & 1 RTX 4090 GPU & 90--160 \\
Complete seed-0 suite & 96 & 1 RTX 4090 GPU & 55--130 \\
Content/style probing diagnostic & 16 & 1 RTX 4090 GPU & 1--3 \\
Parameter-matched adapter diagnostic & 3 & 1 RTX 4090 GPU & 2--5 \\
GRL subject-confusion diagnostic & 15--30 & 1 RTX 4090 GPU & 3--8 \\
Compute and memory overhead benchmark & 2 one-epoch runs & 1 RTX 4090 GPU & $<0.1$ \\
\bottomrule
\end{tabular}}
\end{table}

\paragraph{Dataset licenses and usage terms.}
\label{app:asset_licenses}
Table~\ref{tab:app_dataset_licenses} summarizes the source and usage terms for the datasets in the complete evaluation suite. We use each dataset under its original terms, cite the original sources where available, and do not redistribute raw datasets or pretrained backbone checkpoints in the code release. Non-commercial, no-derivatives, and academic-only terms are reported as such and should not be interpreted as open commercial licenses. For datasets where no standard license name could be identified from the official source metadata, we report that status explicitly.

\footnotesize
\setlength{\tabcolsep}{3.0pt}
\begin{longtable}{@{}p{0.16\textwidth}p{0.37\textwidth}p{0.39\textwidth}@{}}
\caption{\normalsize Dataset sources and usage terms for the complete evaluation suite.}
\label{tab:app_dataset_licenses}\\
\toprule
\textbf{Dataset} & \textbf{Source} & \textbf{License / usage terms} \\
\midrule
\endfirsthead
\toprule
\textbf{Dataset} & \textbf{Source} & \textbf{License / usage terms} \\
\midrule
\endhead
\bottomrule
\endfoot
BCIC-IV-2a & BNCI Horizon 2020 dataset 001-2014, BCI Competition IV dataset 2a; \url{https://bnci-horizon-2020.eu/database/data-sets} & CC BY-ND 4.0. \\
SEED & SJTU BCMI Lab SEED collection; \url{https://bcmi.sjtu.edu.cn/home/seed/} & Application/signature required; academic research only; no commercial use. \\
SEED-IV & SJTU BCMI Lab SEED-IV collection; \url{https://bcmi.sjtu.edu.cn/home/seed/} & Application/signature required; academic research only; no commercial use. \\
SEED-V & SJTU BCMI Lab SEED-V collection; \url{https://bcmi.sjtu.edu.cn/home/seed/} & Application/signature required; academic research only; no commercial use. \\
FACED & FACED dataset, Synapse project syn50614194; \url{https://www.nature.com/articles/s41597-023-02650-w} & CC-BY metadata; the Scientific Data article is CC BY 4.0, with download terms governed by the Synapse page. \\
MI-KoreaU & Lee2019/OpenBMI, GigaDB DOI 10.5524/100542; \url{https://moabb.neurotechx.com/docs/generated/moabb.datasets.Lee2019_MI.html} & GPL 3.0 according to MOABB metadata. \\
MentalArith. & PhysioNet EEG During Mental Arithmetic Tasks; \url{https://physionet.org/content/eegmat/} & Open Data Commons Attribution License v1.0. \\
PhysioNet-MI & PhysioNet EEG Motor Movement/Imagery Dataset; \url{https://www.physionet.org/content/eegmmidb/} & Open Data Commons Attribution License v1.0. \\
Stieger2021-UD & Stieger et al. continuous SMR BCI learning dataset, up/down task split, figshare DOI 10.6084/m9.figshare.13123148.v1; \url{https://moabb.neurotechx.com/docs/generated/moabb.datasets.Stieger2021.html} & CC BY-NC 4.0. \\
Mumtaz2016 & figshare MDD Patients and Healthy Controls EEG Data, DOI 10.6084/m9.figshare.4244171; \url{https://figshare.com/articles/dataset/EEG_Data_New/4244171} & CC BY 4.0. \\
BCIC-Speech & 2020 BCI Competition Track 3 imagined-speech dataset and associated OSF/NeMAR-style metadata; \url{https://brain.korea.ac.kr/bci2020/competition.php} & CC BY 4.0 according to NeMAR/EEGDash-style metadata for the matched speech-decoding dataset. \\
ISRUC-S1 & ISRUC-Sleep dataset, Subgroup 1; \url{https://sleeptight.isr.uc.pt/} & Public/open-access dataset; no explicit standard license name identified in the official/source metadata. \\
\end{longtable}
\normalsize

\section{Dataset Details}
\label{app:dataset_details}

\subsection{Dataset Statistics}
\label{app:dataset_stats}

Table~\ref{tab:dataset_splits} summarizes each dataset's standardized local sample tensor shape, number of classes, task type, and subject-disjoint split statistics. Shapes are listed before model-specific channel selection or backbone-specific reshaping.

\scriptsize
\setlength{\tabcolsep}{2.0pt}
\begin{longtable}{@{}p{0.135\textwidth}p{0.135\textwidth}p{0.055\textwidth}p{0.195\textwidth}p{0.095\textwidth}p{0.095\textwidth}p{0.095\textwidth}p{0.095\textwidth}@{}}
\caption{\normalsize Dataset shapes, class counts, task types, and split statistics used in the reported experiments. Each dataset is shown with subject counts and sample counts in separate rows.}
\label{tab:dataset_splits}\\
\toprule
\textbf{Dataset} & \textbf{Local shape} & \textbf{\# Cls.} & \textbf{Task type} & \textbf{Statistic} & \textbf{Train} & \textbf{Val.} & \textbf{Test} \\
\midrule
\endfirsthead
\toprule
\textbf{Dataset} & \textbf{Local shape} & \textbf{\# Cls.} & \textbf{Task type} & \textbf{Statistic} & \textbf{Train} & \textbf{Val.} & \textbf{Test} \\
\midrule
\endhead
\bottomrule
\endfoot
\multirow{2}{=}{BCIC-IV-2a} & \multirow{2}{=}{\((22,4,200)\)} & \multirow{2}{=}{4} & \multirow{2}{=}{Motor imagery (multi.)} & Subjects & 5 & 2 & 2 \\
 & & & & Samples & 2784 & 1152 & 1152 \\
\addlinespace[1pt]
\multirow{2}{=}{SEED} & \multirow{2}{=}{\((60,4,200)\)} & \multirow{2}{=}{2} & \multirow{2}{=}{Emotion (binary)} & Subjects & 10 & 2 & 3 \\
 & & & & Samples & 2700 & 540 & 810 \\
\addlinespace[1pt]
\multirow{2}{=}{SEED-IV} & \multirow{2}{=}{\((60,4,200)\)} & \multirow{2}{=}{4} & \multirow{2}{=}{Emotion (multi.)} & Subjects & 10 & 2 & 3 \\
 & & & & Samples & 25050 & 5010 & 7515 \\
\addlinespace[1pt]
\multirow{2}{=}{SEED-V} & \multirow{2}{=}{\((62,1,200)\)} & \multirow{2}{=}{5} & \multirow{2}{=}{Emotion (multi.)} & Subjects & 5 & 5 & 5 \\
 & & & & Samples & 36795 & 36795 & 36795 \\
\addlinespace[1pt]
\multirow{2}{=}{FACED} & \multirow{2}{=}{\((32,10,200)\)} & \multirow{2}{=}{9} & \multirow{2}{=}{Emotion (multi.)} & Subjects & 80 & 20 & 23 \\
 & & & & Samples & 6720 & 1680 & 1932 \\
\addlinespace[1pt]
\multirow{2}{=}{MI-KoreaU} & \multirow{2}{=}{\((62,4,200)\)} & \multirow{2}{=}{2} & \multirow{2}{=}{Motor imagery (binary)} & Subjects & 42 & 5 & 7 \\
 & & & & Samples & 16800 & 2000 & 2800 \\
\addlinespace[1pt]
\multirow{2}{=}{MentalArith.} & \multirow{2}{=}{\((20,5,200)\)} & \multirow{2}{=}{2} & \multirow{2}{=}{Workload assessment (binary)} & Subjects & 26 & 5 & 5 \\
 & & & & Samples & 1247 & 240 & 220 \\
\addlinespace[1pt]
\multirow{2}{=}{PhysioNet-MI} & \multirow{2}{=}{\((64,4,200)\)} & \multirow{2}{=}{4} & \multirow{2}{=}{Motor imagery (multi.)} & Subjects & 70 & 19 & 20 \\
 & & & & Samples & 6300 & 1734 & 1803 \\
\addlinespace[1pt]
\multirow{2}{=}{Stieger2021-UD} & \multirow{2}{=}{\((60,4,200)\)} & \multirow{2}{=}{2} & \multirow{2}{=}{Motor imagery (binary)} & Subjects & 40 & 9 & 9 \\
 & & & & Samples & 89982 & 17776 & 19574 \\
\addlinespace[1pt]
\multirow{2}{=}{Mumtaz2016} & \multirow{2}{=}{\((19,5,200)\)} & \multirow{2}{=}{2} & \multirow{2}{=}{MDD classification (binary)} & Subjects & 43 & 9 & 10 \\
 & & & & Samples & 4891 & 1041 & 1151 \\
\addlinespace[1pt]
\multirow{2}{=}{BCIC-Speech} & \multirow{2}{=}{\((60,3,200)\)} & \multirow{2}{=}{3} & \multirow{2}{=}{Speech decoding (multi.)} & Subjects & 9 & 3 & 3 \\
 & & & & Samples & 4050 & 1350 & 1350 \\
\addlinespace[1pt]
\multirow{2}{=}{ISRUC-S1} & \multirow{2}{=}{\((20,6,6000)\)} & \multirow{2}{=}{5} & \multirow{2}{=}{Sleep staging (multi.)} & Subjects & 80 & 10 & 10 \\
 & & & & Samples & 3559 & 468 & 435 \\
\end{longtable}
\normalsize

For most datasets, local sample shapes follow \((\text{channels}, \text{temporal segments}, \text{samples per segment})\). ISRUC is stored as sleep-epoch sequences with shape \((\text{epochs}, \text{channels}, \text{samples})\).

\subsection{Backbone-Specific Preprocessing}
\label{app:backbone_preprocessing}

The downstream experiments reuse the processed EEG inputs from the corresponding foundation-model pipelines without adding a separate raw-signal preprocessing stage for \modelnickname{}. Across datasets, recordings are converted into fixed-duration EEG tensors with labels, subject identifiers, and split keys. Most samples are resampled to 200 Hz and represented as one-second temporal patches; dataset-appropriate filtering and notch removal follow the original dataset or backbone pipeline when raw signals are available.

Backbone-specific preprocessing is limited to the model input interface. CodeBrain and CBraMod consume the shared LMDB or sequence-folder tensors after common amplitude scaling. CSBrain uses the same processed samples but reshapes them to match its tokenization convention; for BCIC-Speech, it uses the corrected 60-channel, three-window sample format. LaBraM additionally maps dataset channel names to its standard channel-index convention for channel embeddings. Thus, the comparisons use the same signal-level preprocessing whenever possible, while differences are limited to tensor shape, channel metadata, resampling rate, or tokenizer requirements.

\section{Implementation Details}
\label{app:implementation_details}

We use the public released checkpoints of CodeBrain, CBraMod, LaBraM, and CSBrain and do not re-train the EFM pretraining stage. Since public checkpoint documentation may not expose subject/session-level pretraining metadata for every downstream dataset, our claims focus on matched downstream adaptation comparisons under the same checkpoint, split, and trainable-backbone policy.

\subsection{Details of Recursive Gaussian Inference}
\label{app:gaussian_details}

Observation~3.1 in the main text focuses on counterpart removal and the resulting branch complementarity. Here we provide the full uncertainty-aware Gaussian update used to implement that residual refinement. The recursive Gaussian disentangler operates on the window-level observation sequence $\{(e_w,\Lambda_w)\}_{w=1}^{W}$ derived from $\bar{\mixedtokens}$. To reduce overlap between the two branches, each update uses a counterpart-removed observation in place of the raw mixed window. For a generic branch update, we write
\begin{equation}
\gamma_w = \sigma(a_w), \qquad
\tilde{e}_w = e_w - \gamma_w r_w, \qquad
\tilde{\Lambda}_w = \left(\Lambda_w^{-1} + (\gamma_w R_w)^{-1}\right)^{-1},
\end{equation}
where $r_w$ and $R_w$ denote the counterpart mean and precision from the opposite branch, and $\gamma_w \in (0,1)$ is a learnable removal gate. This gate controls how much counterpart information is subtracted before the current branch is updated.
All precision terms are implemented as positive diagonal vectors. In particular, the observation precision is parameterized from the predicted log-precision as $\Lambda_w=\mathrm{diag}(\exp(f_{\lambda}(\bar z_w))+\epsilon)$ with a small $\epsilon>0$, and the same positive-diagonal convention is used for the transition and counterpart precisions. The Gaussian inverses below are therefore computed elementwise and remain well defined.

Given the transition prior
\begin{equation}
h_w \mid h_{w-1} \sim \mathcal{N}(T_w h_{w-1}, \Pi_w^{-1}),
\end{equation}
and the counterpart-removed observation likelihood
\begin{equation}
\tilde e_w \mid h_w \sim \mathcal{N}(h_w, \tilde{\Lambda}_w^{-1}),
\end{equation}
the posterior is obtained by Gaussian fusion:
\begin{equation}
p(h_w \mid \tilde{e}_{\leq w}) \propto
\mathcal{N}(h_w; T_w h_{w-1}, \Pi_w^{-1})
\mathcal{N}(\tilde{e}_w; h_w, \tilde{\Lambda}_w^{-1}),
\end{equation}
which yields
\begin{equation}
\Sigma_w = \left(\Pi_w + \tilde{\Lambda}_w\right)^{-1}, \qquad
h_w = \Sigma_w \left(\Pi_w T_w h_{w-1} + \tilde{\Lambda}_w \tilde{e}_w\right).
\end{equation}
This is the detailed Gaussian-fusion derivation referenced in Observation~3.1 of the main text: the posterior remains Gaussian, and its mean is a precision-weighted combination of temporal prediction and counterpart-removed observation.

The style branch uses a static transition $T_w=I$, while the content branch uses the dynamic transition
\begin{equation}
T_w = G_w = \sum_{n=1}^{N} \alpha_{w,n} G_n, \qquad
\boldsymbol{\alpha}_w = \mathrm{Softmax}(g(h_{w-1})),
\end{equation}
where $\{G_n\}_{n=1}^{N}$ are learnable transition bases. The full recursive procedure alternates the two branches in four steps:
\begin{equation}
\left(s_w^{(0)}, P_w^{(0)}\right) = \mathcal{G}_{\mathrm{sta}}(e_w,\Lambda_w),
\end{equation}
\begin{equation}
\left(c_w^{(0)}, Q_w^{(0)}\right) = \mathcal{G}_{\mathrm{dyn}}(e_w,\Lambda_w; s_w^{(0)}, P_w^{(0)}),
\end{equation}
\begin{equation}
\left(s_w, P_w\right) = \mathcal{G}_{\mathrm{sta}}(e_w,\Lambda_w; c_w^{(0)}, Q_w^{(0)}),
\end{equation}
\begin{equation}
\left(c_w, Q_w\right) = \mathcal{G}_{\mathrm{dyn}}(e_w,\Lambda_w; s_w, P_w),
\end{equation}
where the first two steps produce coarse factor estimates and the latter two refine them after counterpart removal. The resulting \textcolor{contentwine}{$\mathbf{C}^{\mathrm{win}}=[c_1,\dots,c_W]$} and \textcolor{styleblue}{$\mathbf{S}^{\mathrm{win}}=[s_1,\dots,s_W]$} are then reinjected into the token grid to form \contentprose{} and \styleprose{} in the main text.

For the reconstruction term in Section~\ref{sec:external_constraint}, we use a lightweight FiLM projection head. Given the sample-level style summary $\mathbf{s}^{\ast}$, the FiLM head predicts a per-dimension scale and bias
\begin{equation}
\lambda, b = g_{\mathrm{film}}(\mathbf{s}^{\ast}),
\end{equation}
and reconstructs the mixed window representation through
\begin{equation}
\hat{z}_w = \lambda \odot c_w + b.
\end{equation}
This implements a simple subject-conditioned linear regression, where both regression parameters are generated from the style embedding.
Because the reconstruction head is intentionally lightweight, it encourages a complementary information split: the content token $c_w$ carries the main content information, while the style embedding $\mathbf{s}^{\ast}$ modulates reconstruction through the FiLM scale and bias.

\subsection{Representation Geometry Metric}
\label{app:representation_geometry_metric}

The visualization analysis in Section~\ref{sec:analysis_visualization} uses a post-hoc within-subject class separation score. This metric is used only for representation analysis and is not part of the training objective. We compute it for the before-module mixed representation and the content-branch representation produced by \modelnickname{}.

For each dataset and representation type, let $e_i$ denote the sample-level representation of trial $i$, with subject label $s_i$ and task label $y_i$. We first remove the subject-specific mean:
\begin{equation}
\tilde{e}_i = e_i - \frac{1}{|\mathcal{I}_{s_i}|}\sum_{j \in \mathcal{I}_{s_i}} e_j,
\end{equation}
where $\mathcal{I}_{s_i}$ is the set of samples from subject $s_i$. The centered representations are then standardized over the analysis set and projected with PCA to at most 50 dimensions, using $\min(50, N-1, D)$ components when the sample count or feature dimension is smaller.

For each valid subject with at least two observed task classes, we compute the centroid $\mu_{s,c}$ of each class $c$ and define
\begin{equation}
B_s = \frac{1}{|\mathcal{P}_s|}\sum_{(c_a,c_b)\in\mathcal{P}_s}
\left\|\mu_{s,c_a}-\mu_{s,c_b}\right\|_2,
\end{equation}
where $\mathcal{P}_s$ contains all unordered pairs of classes observed for subject $s$. The within-class dispersion is
\begin{equation}
W_s = \frac{1}{|\mathcal{I}_s|}\sum_{i\in\mathcal{I}_s}
\left\|z_i-\mu_{s,y_i}\right\|_2,
\end{equation}
where $z_i$ is the PCA-projected representation. The subject-level score is
\begin{equation}
R_s = \frac{B_s}{W_s + 10^{-8}},
\end{equation}
and the final dataset-level score is the average of $R_s$ over valid subjects. Larger values therefore indicate that task-class centroids are more separated relative to within-class dispersion after subject-level offsets have been removed. The t-SNE panels use the same centered, standardized, and PCA-projected representations, with balanced subsampling for visualization when needed.

\subsection{Training Hyperparameters}
\label{app:training_hparams}

Hyperparameters and checkpoints are selected only on the validation split associated with each subject-disjoint partition. ``Representative'' denotes a backbone-level default together with the small dataset-level ranges shown below. For a new EFM, we start from the default configuration and tune the loss weights over the same small validation-only range. The released code includes configuration files for each backbone and dataset. Table~\ref{tab:general_hparams} and Table~\ref{tab:backbone_hparams} summarize the paper-facing hyperparameter settings used in the reported experiments. Across backbones, \modelnickname{} receives the pretrained token grid with shape $(B,M,W,D)$, performs Gaussian inference on channel-pooled windows with shape $(B,W,D)$, reinjects the resulting content/style corrections into the token grid, and feeds only the content branch to the prediction head.

\begin{table}[!t]
\centering
\normalsize
\renewcommand{\arraystretch}{1.05}
\caption{General hyperparameter settings used for disentangled fine-tuning.}
\label{tab:general_hparams}
\setlength{\tabcolsep}{5pt}
\begin{tabular}{@{}ll@{}}
\toprule
\textbf{Hyperparameter} & \textbf{Value / policy} \\
\midrule
Optimizer & AdamW \\
Random seed & \begin{tabular}[t]{@{}l@{}}0--2 for Table~\ref{tab:main_result}.\\ Complete results are reported under seed 0.\end{tabular} \\
Weight decay & 0.05 \\
Dropout & 0.1 \\
Label smoothing & 0.1 \\
Task loss weight & 1.0 \\
Adapter hidden dimension & 200 \\
Gaussian transition components & 16 \\
Remove-gate initialization & -2.0 \\
Contrastive temperature & 0.07 \\
Variance-ratio target & 2.0 \\
\bottomrule
\end{tabular}
\end{table}

\begin{table}[!t]
\centering
\small
\renewcommand{\arraystretch}{1.05}
\caption{Representative backbone-specific optimization and loss-weight settings. The four $\lambda$ columns correspond directly to Eq.~\ref{eq:total_loss}.}
\label{tab:backbone_hparams}
\setlength{\tabcolsep}{3pt}
\begin{tabular}{@{}lccccccc@{}}
\toprule
\textbf{Backbone} & \textbf{Learning rate} & \textbf{Batch size} & \textbf{Epochs} & $\boldsymbol{\lambda_{\mathrm{ord}}}$ & $\boldsymbol{\lambda_{\mathrm{sty}}}$ & $\boldsymbol{\lambda_{\mathrm{con}}}$ & $\boldsymbol{\lambda_{\mathrm{rec}}}$ \\
\midrule
CodeBrain & $1\times10^{-4}$ & 64 & 20 & 0.005 & 0.001 & 0.02 & 0.01 \\
CBraMod & $1\times10^{-4}$ & 64 & 20 & 0.1 & 0.01--0.02 & 0.2 & 0.02 \\
LaBraM & $1\times10^{-4}$ & 64 & 20 & 0.1 & 0.02 & 0.2 & 0.02 \\
CSBrain & $1\times10^{-4}$ & 64 & 20 & 0.1 & 0.01 & 0.01 & 0.02 \\
\bottomrule
\end{tabular}
\end{table}

\subsection{Parameter Overhead}
\label{app:parameter_overhead}

\modelnickname{} reuses the pretrained backbone and downstream prediction head from the matched fine-tuning baseline, and only adds the post-backbone disentanglement adapter. To quantify the overhead, Table~\ref{tab:parameter_overhead} reports both the added adapter parameters and their ratio to the original fine-tuning model size, where the original model includes the pretrained backbone and the downstream prediction head. Across these settings, the active adapter adds 2.07M parameters and accounts for 1.55\%--10.81\% of the original model size. The percentage varies because the downstream prediction head changes with the task interface and label space, while the CBraMod backbone and the adapter size remain fixed.

\begin{table}[!t]
\centering
\small
\renewcommand{\arraystretch}{1.05}
\caption{Parameter overhead of \modelnickname{} on the six representative CBraMod settings. The original model is the matched vanilla fine-tuning model, including the pretrained backbone and downstream prediction head.}
\label{tab:parameter_overhead}
\setlength{\tabcolsep}{4pt}
\begin{tabular}{@{}lccc@{}}
\toprule
\textbf{Dataset} & \textbf{Original model} & \textbf{Added adapter} & \textbf{Overhead} \\
\midrule
BCIC-IV-2a & 19.13M & 2.07M & 10.81\% \\
FACED & 133.29M & 2.07M & 1.55\% \\
ISRUC-S1 & 26.47M & 2.07M & 7.81\% \\
MentalArith. & 25.09M & 2.07M & 8.24\% \\
BCIC-Speech & 26.61M & 2.07M & 7.77\% \\
Mumtaz2016 & 24.09M & 2.07M & 8.58\% \\
\bottomrule
\end{tabular}
\end{table}

\subsection{Compute and Memory Overhead}
\label{app:runtime_memory_overhead}

\modelnickname{} adds overhead through the post-backbone adapter and auxiliary training objectives. At inference time, subject identities are not required, and prediction is performed through the content branch. Table~\ref{tab:runtime_memory_overhead} summarizes the measured parameter, memory, and one-step training FLOP overhead on a representative CBraMod setting.

\begin{table}[!t]
\centering
\small
\renewcommand{\arraystretch}{1.05}
\caption{Compute and memory overhead diagnostic for a representative CBraMod setting. FLOPs are measured for one training step.}
\label{tab:runtime_memory_overhead}
\setlength{\tabcolsep}{4pt}
\resizebox{\textwidth}{!}{%
\begin{tabular}{@{}lcccc@{}}
\toprule
\textbf{Setting} & \textbf{Trainable params} & \textbf{Peak allocated memory} & \textbf{Peak reserved memory} & \textbf{One-step train FLOPs} \\
\midrule
Vanilla & 19.13M & 879.7 MiB & 1216.0 MiB & 85.54G \\
\modelnickname{} & 21.44M & 970.1 MiB & 1300.0 MiB & 93.86G \\
Overhead & +2.32M (+12.11\%) & +90.4 MiB (+10.28\%) & +84.0 MiB (+6.91\%) & +8.33G (+9.74\%) \\
\bottomrule
\end{tabular}}
\end{table}

\section{Complete Main Results}
\label{app:complete_main_results}

\makeatletter
\setlength{\@fptop}{0pt}
\setlength{\@fpsep}{8pt}
\setlength{\@fpbot}{0pt plus 1fil}
\makeatother

To assess broader coverage, this section reports seed-0 results across the complete evaluation suite. These results are intended as coverage evidence, while Table~\ref{tab:main_result} remains the primary multi-seed quantitative comparison. Tables~\ref{tab:app_main_codebrain}--\ref{tab:app_main_csbrain} keep one backbone fixed and report all recorded metrics for vanilla fine-tuning and the \modelnickname{} variant, so the comparison remains within the same pretrained representation.

CodeBrain shows clear improvements on BCIC-IV-2a, FACED, MentalArith., BCIC-Speech, and ISRUC-S1, covering motor imagery, emotion recognition, cognitive-state decoding, speech decoding, and sleep staging. The remaining lower secondary metrics are generally smaller than the gains on the representative datasets.

\begin{table}[!t]
\centering
\scriptsize
\renewcommand{\arraystretch}{1.04}
\caption{Complete main results for CodeBrain across the evaluation suite and recorded metrics. Positive deltas are shaded orange, negative deltas are shaded blue, and the better value between Vanilla and \modelnickname{} is bold.}
\label{tab:app_main_codebrain}
\setlength{\tabcolsep}{3.0pt}
\resizebox{0.997\textwidth}{!}{%
\begin{tabular}{@{}llccccccccc@{}}
\toprule
\textbf{Setting} & \multicolumn{3}{c}{\textbf{BCIC-IV-2a}} & \multicolumn{3}{c}{\textbf{SEED}} & \multicolumn{3}{c}{\textbf{SEED-IV}} \\
\cmidrule(lr){2-4} \cmidrule(lr){5-7} \cmidrule(lr){8-10}
 & B-ACC & Kappa & F1 & B-ACC & \mbox{ROC-AUC} & \mbox{PR-AUC} & B-ACC & Kappa & F1 \\
\midrule
Vanilla & 0.3602 & 0.1470 & 0.2815 & 0.6827 & 0.7499 & 0.7559 & 0.3403 & 0.1151 & 0.2982 \\
\modelnickname{} & \bestperf{0.4740} & \bestperf{0.2986} & \bestperf{0.4523} & \bestperf{0.6914} & 0.7750 & 0.8068 & \bestperf{0.3515} & \bestperf{0.1277} & \bestperf{0.3025} \\
$\Delta$ & \posdelta{+0.1138} & \posdelta{+0.1516} & \posdelta{+0.1708} & \posdelta{+0.0087} & \posdelta{+0.0251} & \posdelta{+0.0508} & \posdelta{+0.0112} & \posdelta{+0.0126} & \posdelta{+0.0043} \\
\midrule
\textbf{Setting} & \multicolumn{3}{c}{\textbf{SEED-V}} & \multicolumn{3}{c}{\textbf{FACED}} & \multicolumn{3}{c}{\textbf{MI-KoreaU}} \\
\cmidrule(lr){2-4} \cmidrule(lr){5-7} \cmidrule(lr){8-10}
 & B-ACC & Kappa & F1 & B-ACC & Kappa & F1 & B-ACC & \mbox{ROC-AUC} & \mbox{PR-AUC} \\
\midrule
Vanilla & 0.2672 & 0.0835 & \bestperf{0.2651} & 0.4866 & 0.4212 & 0.4864 & 0.7761 & 0.8324 & 0.7926 \\
\modelnickname{} & \bestperf{0.2743} & \bestperf{0.0919} & 0.2618 & \bestperf{0.5784} & \bestperf{0.5217} & \bestperf{0.5761} & \bestperf{0.8096} & 0.8683 & 0.8707 \\
$\Delta$ & \posdelta{+0.0071} & \posdelta{+0.0084} & \negdelta{-0.0033} & \posdelta{+0.0918} & \posdelta{+0.1005} & \posdelta{+0.0897} & \posdelta{+0.0335} & \posdelta{+0.0359} & \posdelta{+0.0781} \\
\midrule
\textbf{Setting} & \multicolumn{3}{c}{\textbf{MentalArith.}} & \multicolumn{3}{c}{\textbf{PhysioNet}} & \multicolumn{3}{c}{\textbf{Stieger2021-UD}} \\
\cmidrule(lr){2-4} \cmidrule(lr){5-7} \cmidrule(lr){8-10}
 & B-ACC & \mbox{ROC-AUC} & \mbox{PR-AUC} & B-ACC & \mbox{ROC-AUC} & \mbox{PR-AUC} & B-ACC & Kappa & F1 \\
\midrule
Vanilla & 0.6938 & 0.7248 & 0.5368 & 0.6157 & 0.4875 & 0.6177 & 0.7566 & \bestperf{0.8545} & \bestperf{0.8623} \\
\modelnickname{} & \bestperf{0.7583} & \bestperf{0.8266} & \bestperf{0.6781} & \bestperf{0.6228} & \bestperf{0.4971} & \bestperf{0.6256} & \bestperf{0.7632} & 0.8422 & 0.8518 \\
$\Delta$ & \posdelta{+0.0645} & \posdelta{+0.1018} & \posdelta{+0.1413} & \posdelta{+0.0071} & \posdelta{+0.0096} & \posdelta{+0.0079} & \posdelta{+0.0066} & \negdelta{-0.0123} & \negdelta{-0.0105} \\
\bottomrule
\end{tabular}}
\vspace{3pt}
\setlength{\tabcolsep}{3.0pt}
\resizebox{0.997\textwidth}{!}{%
\begin{tabular}{@{}llccccccccc@{}}
\toprule
\textbf{Setting} & \multicolumn{3}{c}{\textbf{Mumtaz2016}} & \multicolumn{3}{c}{\textbf{BCIC-Speech}} & \multicolumn{3}{c}{\textbf{ISRUC-S1}} \\
\cmidrule(lr){2-4} \cmidrule(lr){5-7} \cmidrule(lr){8-10}
 & B-ACC & \mbox{ROC-AUC} & \mbox{PR-AUC} & B-ACC & Kappa & F1 & B-ACC & Kappa & F1 \\
\midrule
Vanilla & 0.9043 & 0.9732 & \bestperf{0.9796} & 0.4916 & 0.2373 & 0.4865 & 0.7698 & 0.7175 & 0.7783 \\
\modelnickname{} & \bestperf{0.9110} & \bestperf{0.9742} & 0.9784 & \bestperf{0.5116} & \bestperf{0.2673} & \bestperf{0.5119} & \bestperf{0.7703} & \bestperf{0.7437} & \bestperf{0.7985} \\
$\Delta$ & \posdelta{+0.0067} & \posdelta{+0.0010} & \negdelta{-0.0012} & \posdelta{+0.0200} & \posdelta{+0.0300} & \posdelta{+0.0254} & \posdelta{+0.0005} & \posdelta{+0.0262} & \posdelta{+0.0202} \\
\bottomrule
\end{tabular}}
\end{table}
\clearpage

CBraMod shows the strongest breadth pattern in the seed-0 records, with clear gains across several task families and a small number of metric-level regressions.

\begin{table}[!t]
\centering
\scriptsize
\renewcommand{\arraystretch}{1.04}
\caption{Complete main results for CBraMod across the evaluation suite and recorded metrics. Positive deltas are shaded orange, negative deltas are shaded blue, and the better value between Vanilla and \modelnickname{} is bold.}
\label{tab:app_main_cbramod}
\setlength{\tabcolsep}{3.0pt}
\resizebox{0.997\textwidth}{!}{%
\begin{tabular}{@{}llccccccccc@{}}
\toprule
\textbf{Setting} & \multicolumn{3}{c}{\textbf{BCIC-IV-2a}} & \multicolumn{3}{c}{\textbf{SEED}} & \multicolumn{3}{c}{\textbf{SEED-IV}} \\
\cmidrule(lr){2-4} \cmidrule(lr){5-7} \cmidrule(lr){8-10}
 & B-ACC & Kappa & F1 & B-ACC & \mbox{ROC-AUC} & \mbox{PR-AUC} & B-ACC & Kappa & F1 \\
\midrule
Vanilla & 0.4870 & 0.3160 & 0.4650 & 0.7383 & 0.8346 & 0.8489 & 0.3773 & 0.1601 & 0.3301 \\
\modelnickname{} & \bestperf{0.5981} & \bestperf{0.4641} & \bestperf{0.5954} & \bestperf{0.7877} & \bestperf{0.8589} & \bestperf{0.8647} & \bestperf{0.3989} & \bestperf{0.1897} & \bestperf{0.3444} \\
$\Delta$ & \posdelta{+0.1111} & \posdelta{+0.1481} & \posdelta{+0.1304} & \posdelta{+0.0494} & \posdelta{+0.0243} & \posdelta{+0.0158} & \posdelta{+0.0216} & \posdelta{+0.0296} & \posdelta{+0.0143} \\
\midrule
\textbf{Setting} & \multicolumn{3}{c}{\textbf{SEED-V}} & \multicolumn{3}{c}{\textbf{FACED}} & \multicolumn{3}{c}{\textbf{MI-KoreaU}} \\
\cmidrule(lr){2-4} \cmidrule(lr){5-7} \cmidrule(lr){8-10}
 & B-ACC & Kappa & F1 & B-ACC & Kappa & F1 & B-ACC & \mbox{ROC-AUC} & \mbox{PR-AUC} \\
\midrule
Vanilla & 0.2465 & 0.0600 & 0.2490 & 0.5451 & 0.4830 & 0.5436 & 0.8343 & 0.9164 & 0.9181 \\
\modelnickname{} & \bestperf{0.2734} & \bestperf{0.0970} & \bestperf{0.2746} & \bestperf{0.6126} & \bestperf{0.5605} & \bestperf{0.6128} & \bestperf{0.8450} & \bestperf{0.9170} & \bestperf{0.9221} \\
$\Delta$ & \posdelta{+0.0269} & \posdelta{+0.0370} & \posdelta{+0.0256} & \posdelta{+0.0675} & \posdelta{+0.0775} & \posdelta{+0.0692} & \posdelta{+0.0107} & \posdelta{+0.0006} & \posdelta{+0.0040} \\
\midrule
\textbf{Setting} & \multicolumn{3}{c}{\textbf{MentalArith.}} & \multicolumn{3}{c}{\textbf{PhysioNet}} & \multicolumn{3}{c}{\textbf{Stieger2021-UD}} \\
\cmidrule(lr){2-4} \cmidrule(lr){5-7} \cmidrule(lr){8-10}
 & B-ACC & \mbox{ROC-AUC} & \mbox{PR-AUC} & B-ACC & \mbox{ROC-AUC} & \mbox{PR-AUC} & B-ACC & Kappa & F1 \\
\midrule
Vanilla & 0.6521 & 0.7408 & 0.5609 & 0.6095 & 0.4794 & 0.6133 & 0.7683 & \bestperf{0.8591} & \bestperf{0.8666} \\
\modelnickname{} & \bestperf{0.7260} & \bestperf{0.8376} & \bestperf{0.6743} & \bestperf{0.6424} & \bestperf{0.5230} & \bestperf{0.6413} & \bestperf{0.7730} & 0.8575 & 0.8633 \\
$\Delta$ & \posdelta{+0.0739} & \posdelta{+0.0968} & \posdelta{+0.1134} & \posdelta{+0.0329} & \posdelta{+0.0436} & \posdelta{+0.0280} & \posdelta{+0.0047} & \negdelta{-0.0016} & \negdelta{-0.0033} \\
\bottomrule
\end{tabular}}
\vspace{3pt}
\setlength{\tabcolsep}{3.0pt}
\resizebox{0.997\textwidth}{!}{%
\begin{tabular}{@{}llccccccccc@{}}
\toprule
\textbf{Setting} & \multicolumn{3}{c}{\textbf{Mumtaz2016}} & \multicolumn{3}{c}{\textbf{BCIC-Speech}} & \multicolumn{3}{c}{\textbf{ISRUC-S1}} \\
\cmidrule(lr){2-4} \cmidrule(lr){5-7} \cmidrule(lr){8-10}
 & B-ACC & \mbox{ROC-AUC} & \mbox{PR-AUC} & B-ACC & Kappa & F1 & B-ACC & Kappa & F1 \\
\midrule
Vanilla & 0.9423 & 0.9902 & 0.9902 & 0.5053 & 0.2580 & 0.5060 & 0.7889 & \bestperf{0.7598} & \bestperf{0.8123} \\
\modelnickname{} & \bestperf{0.9644} & \bestperf{0.9943} & \bestperf{0.9958} & \bestperf{0.5116} & \bestperf{0.2673} & \bestperf{0.5124} & \bestperf{0.7925} & 0.7423 & 0.8049 \\
$\Delta$ & \posdelta{+0.0221} & \posdelta{+0.0041} & \posdelta{+0.0056} & \posdelta{+0.0063} & \posdelta{+0.0093} & \posdelta{+0.0064} & \posdelta{+0.0036} & \negdelta{-0.0175} & \negdelta{-0.0074} \\
\bottomrule
\end{tabular}}
\end{table}
\clearpage

LaBraM gives more conservative gains, which is consistent with its stronger vanilla starting point in the main benchmark. Its gains are most visible on BCIC-IV-2a, FACED, MI-KoreaU, PhysioNet, BCIC-Speech, and ISRUC-S1, while a few probabilistic metrics remain challenging under this backbone.

\begin{table}[!t]
\centering
\scriptsize
\renewcommand{\arraystretch}{1.04}
\caption{Complete main results for LaBraM across the evaluation suite and recorded metrics. Positive deltas are shaded orange, negative deltas are shaded blue, and the better value between Vanilla and \modelnickname{} is bold.}
\label{tab:app_main_labram}
\setlength{\tabcolsep}{3.0pt}
\resizebox{0.997\textwidth}{!}{%
\begin{tabular}{@{}llccccccccc@{}}
\toprule
\textbf{Setting} & \multicolumn{3}{c}{\textbf{BCIC-IV-2a}} & \multicolumn{3}{c}{\textbf{SEED}} & \multicolumn{3}{c}{\textbf{SEED-IV}} \\
\cmidrule(lr){2-4} \cmidrule(lr){5-7} \cmidrule(lr){8-10}
 & B-ACC & Kappa & F1 & B-ACC & \mbox{ROC-AUC} & \mbox{PR-AUC} & B-ACC & Kappa & F1 \\
\midrule
Vanilla & 0.5312 & 0.3750 & 0.5226 & 0.7630 & 0.8261 & 0.8033 & 0.4078 & 0.2008 & 0.3540 \\
\modelnickname{} & \bestperf{0.5703} & \bestperf{0.4271} & \bestperf{0.5554} & \bestperf{0.7667} & \bestperf{0.8288} & \bestperf{0.8154} & \bestperf{0.4132} & \bestperf{0.2086} & \bestperf{0.3584} \\
$\Delta$ & \posdelta{+0.0391} & \posdelta{+0.0521} & \posdelta{+0.0328} & \posdelta{+0.0037} & \posdelta{+0.0027} & \posdelta{+0.0121} & \posdelta{+0.0054} & \posdelta{+0.0078} & \posdelta{+0.0044} \\
\midrule
\textbf{Setting} & \multicolumn{3}{c}{\textbf{SEED-V}} & \multicolumn{3}{c}{\textbf{FACED}} & \multicolumn{3}{c}{\textbf{MI-KoreaU}} \\
\cmidrule(lr){2-4} \cmidrule(lr){5-7} \cmidrule(lr){8-10}
 & B-ACC & Kappa & F1 & B-ACC & Kappa & F1 & B-ACC & \mbox{ROC-AUC} & \mbox{PR-AUC} \\
\midrule
Vanilla & 0.2716 & 0.0904 & 0.2648 & 0.4787 & 0.4108 & 0.4792 & 0.8546 & 0.9186 & 0.9201 \\
\modelnickname{} & \bestperf{0.2825} & \bestperf{0.1017} & \bestperf{0.2761} & \bestperf{0.5030} & \bestperf{0.4380} & \bestperf{0.5060} & \bestperf{0.8707} & \bestperf{0.9427} & \bestperf{0.9436} \\
$\Delta$ & \posdelta{+0.0109} & \posdelta{+0.0113} & \posdelta{+0.0113} & \posdelta{+0.0243} & \posdelta{+0.0272} & \posdelta{+0.0268} & \posdelta{+0.0161} & \posdelta{+0.0241} & \posdelta{+0.0235} \\
\midrule
\textbf{Setting} & \multicolumn{3}{c}{\textbf{MentalArith.}} & \multicolumn{3}{c}{\textbf{PhysioNet}} & \multicolumn{3}{c}{\textbf{Stieger2021-UD}} \\
\cmidrule(lr){2-4} \cmidrule(lr){5-7} \cmidrule(lr){8-10}
 & B-ACC & \mbox{ROC-AUC} & \mbox{PR-AUC} & B-ACC & \mbox{ROC-AUC} & \mbox{PR-AUC} & B-ACC & Kappa & F1 \\
\midrule
Vanilla & 0.7250 & 0.7611 & 0.6092 & 0.6252 & 0.5001 & 0.6253 & 0.7864 & \bestperf{0.8732} & \bestperf{0.8787} \\
\modelnickname{} & \bestperf{0.7312} & \bestperf{0.8191} & \bestperf{0.6904} & \bestperf{0.6495} & \bestperf{0.5326} & \bestperf{0.6514} & \bestperf{0.7921} & 0.8725 & 0.8784 \\
$\Delta$ & \posdelta{+0.0062} & \posdelta{+0.0580} & \posdelta{+0.0812} & \posdelta{+0.0243} & \posdelta{+0.0325} & \posdelta{+0.0261} & \posdelta{+0.0057} & \negdelta{-0.0007} & \negdelta{-0.0003} \\
\bottomrule
\end{tabular}}
\vspace{3pt}
\setlength{\tabcolsep}{3.0pt}
\resizebox{0.997\textwidth}{!}{%
\begin{tabular}{@{}llccccccccc@{}}
\toprule
\textbf{Setting} & \multicolumn{3}{c}{\textbf{Mumtaz2016}} & \multicolumn{3}{c}{\textbf{BCIC-Speech}} & \multicolumn{3}{c}{\textbf{ISRUC-S1}} \\
\cmidrule(lr){2-4} \cmidrule(lr){5-7} \cmidrule(lr){8-10}
 & B-ACC & \mbox{ROC-AUC} & \mbox{PR-AUC} & B-ACC & Kappa & F1 & B-ACC & Kappa & F1 \\
\midrule
Vanilla & 0.9215 & \bestperf{0.9879} & \bestperf{0.9892} & 0.4791 & 0.2187 & 0.4680 & 0.7927 & 0.7566 & 0.8088 \\
\modelnickname{} & \bestperf{0.9262} & 0.9875 & 0.9890 & \bestperf{0.4947} & \bestperf{0.2420} & \bestperf{0.4893} & \bestperf{0.8051} & \bestperf{0.7792} & \bestperf{0.8288} \\
$\Delta$ & \posdelta{+0.0047} & \negdelta{-0.0004} & \negdelta{-0.0002} & \posdelta{+0.0156} & \posdelta{+0.0233} & \posdelta{+0.0213} & \posdelta{+0.0124} & \posdelta{+0.0226} & \posdelta{+0.0200} \\
\bottomrule
\end{tabular}}
\end{table}
\clearpage

CSBrain also benefits from disentangled fine-tuning across a broad set of datasets, with clear gains on BCIC-IV-2a, MI-KoreaU, MentalArith., BCIC-Speech, and ISRUC-S1. Its remaining lower values are limited to isolated secondary metrics or high-baseline binary settings.

\begin{table}[!t]
\centering
\scriptsize
\renewcommand{\arraystretch}{1.04}
\caption{Complete main results for CSBrain across the evaluation suite and recorded metrics. Positive deltas are shaded orange, negative deltas are shaded blue, and the better value between Vanilla and \modelnickname{} is bold.}
\label{tab:app_main_csbrain}
\setlength{\tabcolsep}{3.0pt}
\resizebox{0.997\textwidth}{!}{%
\begin{tabular}{@{}llccccccccc@{}}
\toprule
\textbf{Setting} & \multicolumn{3}{c}{\textbf{BCIC-IV-2a}} & \multicolumn{3}{c}{\textbf{SEED}} & \multicolumn{3}{c}{\textbf{SEED-IV}} \\
\cmidrule(lr){2-4} \cmidrule(lr){5-7} \cmidrule(lr){8-10}
 & B-ACC & Kappa & F1 & B-ACC & \mbox{ROC-AUC} & \mbox{PR-AUC} & B-ACC & Kappa & F1 \\
\midrule
Vanilla & 0.4792 & 0.3056 & 0.4545 & 0.7160 & 0.7920 & \bestperf{0.8223} & 0.3341 & 0.1009 & 0.2590 \\
\modelnickname{} & \bestperf{0.5000} & \bestperf{0.3333} & \bestperf{0.4785} & \bestperf{0.7370} & \bestperf{0.8106} & 0.8109 & \bestperf{0.3357} & \bestperf{0.1090} & \bestperf{0.2840} \\
$\Delta$ & \posdelta{+0.0208} & \posdelta{+0.0278} & \posdelta{+0.0240} & \posdelta{+0.0210} & \posdelta{+0.0185} & \negdelta{-0.0115} & \posdelta{+0.0016} & \posdelta{+0.0080} & \posdelta{+0.0250} \\
\midrule
\textbf{Setting} & \multicolumn{3}{c}{\textbf{SEED-V}} & \multicolumn{3}{c}{\textbf{FACED}} & \multicolumn{3}{c}{\textbf{MI-KoreaU}} \\
\cmidrule(lr){2-4} \cmidrule(lr){5-7} \cmidrule(lr){8-10}
 & B-ACC & Kappa & F1 & B-ACC & Kappa & F1 & B-ACC & \mbox{ROC-AUC} & \mbox{PR-AUC} \\
\midrule
Vanilla & 0.2605 & 0.0785 & 0.2605 & 0.5706 & 0.5143 & 0.5786 & 0.8200 & 0.9045 & 0.9067 \\
\modelnickname{} & \bestperf{0.2699} & \bestperf{0.0887} & \bestperf{0.2636} & \bestperf{0.5790} & \bestperf{0.5230} & \bestperf{0.5830} & \bestperf{0.8407} & \bestperf{0.9188} & \bestperf{0.9196} \\
$\Delta$ & \posdelta{+0.0094} & \posdelta{+0.0102} & \posdelta{+0.0032} & \posdelta{+0.0085} & \posdelta{+0.0087} & \posdelta{+0.0044} & \posdelta{+0.0207} & \posdelta{+0.0143} & \posdelta{+0.0129} \\
\midrule
\textbf{Setting} & \multicolumn{3}{c}{\textbf{MentalArith.}} & \multicolumn{3}{c}{\textbf{PhysioNet}} & \multicolumn{3}{c}{\textbf{Stieger2021-UD}} \\
\cmidrule(lr){2-4} \cmidrule(lr){5-7} \cmidrule(lr){8-10}
 & B-ACC & \mbox{ROC-AUC} & \mbox{PR-AUC} & B-ACC & \mbox{ROC-AUC} & \mbox{PR-AUC} & B-ACC & Kappa & F1 \\
\midrule
Vanilla & 0.5135 & \bestperf{0.7292} & 0.4784 & 0.6273 & 0.5030 & 0.6307 & 0.7425 & 0.8510 & 0.8548 \\
\modelnickname{} & \bestperf{0.6115} & 0.7229 & \bestperf{0.4931} & \bestperf{0.6367} & \bestperf{0.5156} & \bestperf{0.6382} & \bestperf{0.7752} & \bestperf{0.8637} & \bestperf{0.8697} \\
$\Delta$ & \posdelta{+0.0979} & \negdelta{-0.0062} & \posdelta{+0.0147} & \posdelta{+0.0094} & \posdelta{+0.0126} & \posdelta{+0.0075} & \posdelta{+0.0327} & \posdelta{+0.0127} & \posdelta{+0.0149} \\
\bottomrule
\end{tabular}}
\vspace{3pt}
\setlength{\tabcolsep}{3.0pt}
\resizebox{0.997\textwidth}{!}{%
\begin{tabular}{@{}llccccccccc@{}}
\toprule
\textbf{Setting} & \multicolumn{3}{c}{\textbf{Mumtaz2016}} & \multicolumn{3}{c}{\textbf{BCIC-Speech}} & \multicolumn{3}{c}{\textbf{ISRUC-S1}} \\
\cmidrule(lr){2-4} \cmidrule(lr){5-7} \cmidrule(lr){8-10}
 & B-ACC & \mbox{ROC-AUC} & \mbox{PR-AUC} & B-ACC & Kappa & F1 & B-ACC & Kappa & F1 \\
\midrule
Vanilla & \bestperf{0.8995} & 0.9796 & 0.9825 & 0.4907 & 0.2360 & 0.4905 & 0.7698 & 0.7175 & 0.7783 \\
\modelnickname{} & 0.8927 & \bestperf{0.9862} & \bestperf{0.9873} & \bestperf{0.5031} & \bestperf{0.2547} & \bestperf{0.5039} & \bestperf{0.7700} & \bestperf{0.7376} & \bestperf{0.7947} \\
$\Delta$ & \negdelta{-0.0068} & \posdelta{+0.0066} & \posdelta{+0.0048} & \posdelta{+0.0124} & \posdelta{+0.0187} & \posdelta{+0.0134} & \posdelta{+0.0002} & \posdelta{+0.0201} & \posdelta{+0.0164} \\
\bottomrule
\end{tabular}}
\end{table}

Overall, the complete seed-0 records support the broader applicability of \modelnickname{} and also reveal that gains vary with dataset, backbone, and metric choice. The remaining decreases are concentrated in isolated high-baseline or score-calibration-sensitive metrics, suggesting broad improvement in task-oriented decoding across backbones.

\begin{figure}[!t]
\centering
\includegraphics[width=0.98\textwidth]{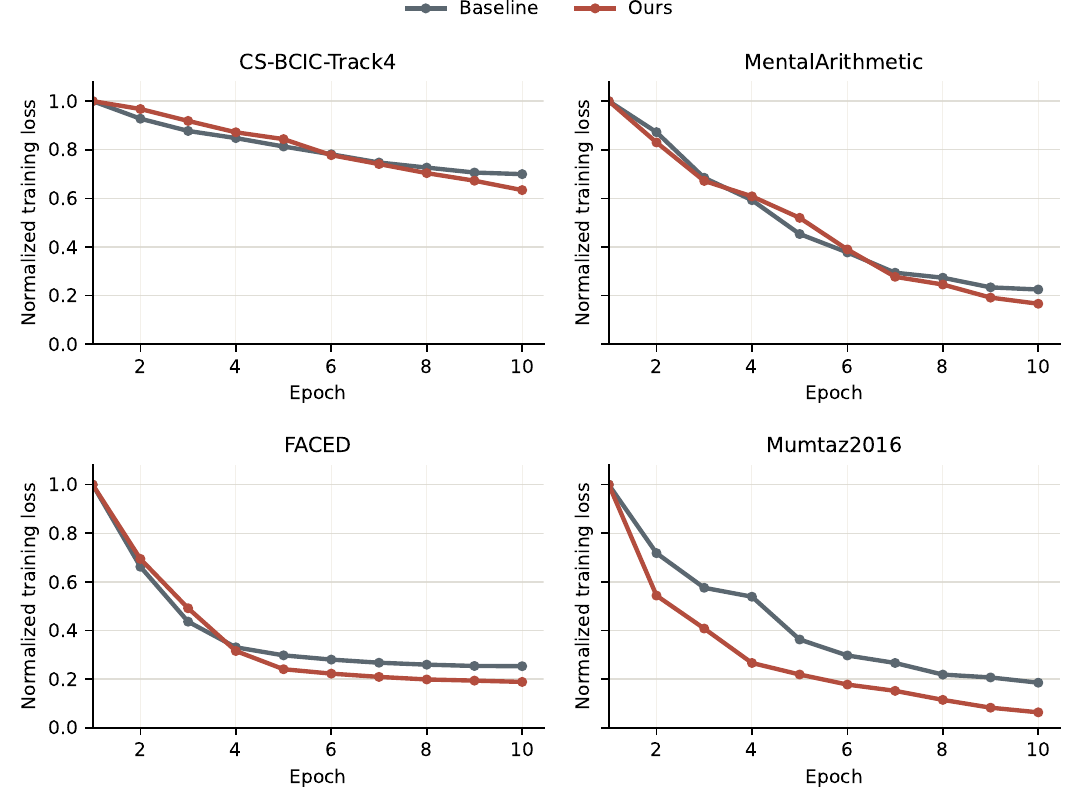}
\caption{\textbf{Normalized training loss curves on CBraMod.} Each curve is divided by its first-epoch loss. The auxiliary constraints in \modelnickname{} provide additional optimization signals and lead to faster normalized loss reduction on the shown datasets.}
\label{fig:app_loss_curves}
\end{figure}

Figure~\ref{fig:app_loss_curves} further compares the training dynamics of vanilla fine-tuning and \modelnickname{} on CBraMod. For each dataset and method, the recorded training loss is normalized by the first-epoch loss, i.e., $\tilde{\mathcal{L}}_t=\mathcal{L}_t/\mathcal{L}_1$. Under this normalized view, \modelnickname{} decreases faster and reaches a lower final loss on the four diagnostic datasets. This pattern is consistent with the role of the auxiliary constraints: subject-aware contrast, temporal-variance shaping, and reconstruction provide additional training signals that guide the adapter toward a more structured content branch during fine-tuning.

\clearpage
\section{Extensive Experiments}
\label{app:extensive_experiments}

\subsection{Content/Style Probing Diagnostic}
\label{app:disentanglement_comparison}

Table~\ref{tab:app_factor_prediction} evaluates the branch outputs immediately before the prediction head. We freeze the trained backbone and disentangler, mean-pool the token grid, and fit one-layer linear probes. Task probes follow the subject-disjoint downstream split; subject probes use held-out trials within a closed-set identity space. Probe checkpoints are selected on the corresponding validation split.

\begin{table}[!t]
\centering
\small
\renewcommand{\arraystretch}{1.05}
\caption{Frozen linear probes on content and style. All values are B-ACC. Task chance is determined by the dataset class count.}
\label{tab:app_factor_prediction}
\setlength{\tabcolsep}{4pt}
\begin{tabular}{@{}llccccc@{}}
\toprule
\textbf{Backbone} & \textbf{Dataset} & \textbf{Task chance} & \textbf{Content task} & \textbf{Style task} & \textbf{Content subject} & \textbf{Style subject} \\
\midrule
CBraMod & FACED & 0.1111 & \bestperf{0.2615} & 0.2458 & 0.9206 & \bestperf{0.9451} \\
CBraMod & BCIC-IV-2a & 0.2500 & \bestperf{0.3409} & 0.3183 & 0.8222 & \bestperf{0.8252} \\
CBraMod & MentalArith. & 0.5000 & \bestperf{0.6705} & 0.6503 & 0.1734 & \bestperf{0.1770} \\
CBraMod & BCIC-Speech & 0.3333 & \bestperf{0.3958} & 0.3849 & 0.3364 & \bestperf{0.3377} \\
CodeBrain & FACED & 0.1111 & \bestperf{0.2684} & 0.2595 & 0.9705 & \bestperf{0.9716} \\
CBraMod & Mumtaz2016 & 0.5000 & \bestperf{0.9759} & 0.9298 & 0.0633 & \bestperf{0.1999} \\
\bottomrule
\end{tabular}
\end{table}

Across all six settings, content yields higher task-probe B-ACC and style yields higher subject-probe B-ACC. On Mumtaz2016, the original/content/style task probes are 0.9224/0.9759/0.9298, and the corresponding subject probes are 0.3720/0.0633/0.1999. These results quantify the functional specialization of the branches.

\subsection{Parameter and Sampler Controls}
\label{app:param_matched_adapter}

The Generic Adapter is a parameter-matched MLP placed between the backbone and classifier. It retains the task loss and uses the same subject-disjoint split, subject-balanced sampler, and backbone update policy as \modelnickname{}.

\begin{table}[!t]
\centering
\small
\renewcommand{\arraystretch}{1.05}
\caption{Stable multi-seed capacity controls. Values are test B-ACC mean$\pm$std.}
\label{tab:app_param_matched_adapter}
\setlength{\tabcolsep}{4pt}
\resizebox{0.82\textwidth}{!}{%
\begin{tabular}{@{}llcc@{}}
\toprule
\textbf{Backbone} & \textbf{Dataset} & \textbf{Generic Adapter} & \textbf{\modelnickname{}} \\
\midrule
CBraMod & BCIC-IV-2a & 0.4806$\pm$0.0403 & \bestperf{0.5579$\pm$0.0310} \\
CBraMod & MentalArith. & 0.6597$\pm$0.0371 & \bestperf{0.7292$\pm$0.0059} \\
LaBraM & BCIC-IV-2a & 0.5483$\pm$0.0067 & \bestperf{0.5671$\pm$0.0032} \\
LaBraM & MentalArith. & 0.6389$\pm$0.0620 & \bestperf{0.7236$\pm$0.0149} \\
\bottomrule
\end{tabular}}
\end{table}

On CBraMod--FACED, Vanilla, a doubled classifier, Vanilla with the subject-balanced sampler, and \modelnickname{} obtain B-ACC of 0.5387, 0.5363, 0.5441, and 0.6005, respectively. The comparison separates classifier capacity and sampling from the full adapter.

\subsection{Subject-Aware and Sequential Baselines}
\label{app:grl_subject_confusion}

Gradient Reversal Layer (GRL) \citep{ganin_domain-adversarial_2016} adds a subject classifier after the EFM tokens and reverses its gradient to learn a subject-confused representation. FHVAE-token \citep{hsu_unsupervised_2017} and DSAE-token \citep{li_disentangled_2018} implement sequential static/dynamic factorization at the same token interface. All methods use the same subject-disjoint splits and samplers.

\begin{table}[!t]
\centering
\small
\renewcommand{\arraystretch}{1.05}
\caption{Multi-seed subject-aware and sequential baselines on FACED. Values are test B-ACC mean$\pm$std.}
\label{tab:app_grl_summary}
\setlength{\tabcolsep}{5pt}
\begin{tabular}{@{}lcccc@{}}
\toprule
\textbf{Backbone} & \textbf{GRL} & \textbf{FHVAE-token} & \textbf{DSAE-token} & \textbf{\modelnickname{}} \\
\midrule
CBraMod & 0.5573$\pm$0.0028 & 0.5883$\pm$0.0052 & 0.5848$\pm$0.0019 & \bestperf{0.6005$\pm$0.0101} \\
LaBraM & 0.5261$\pm$0.0151 & 0.4709$\pm$0.0070 & 0.4828$\pm$0.0031 & \bestperf{0.5409$\pm$0.0086} \\
\bottomrule
\end{tabular}
\end{table}

On BCIC-IV-2a, GRL/FHVAE-token obtain 0.4754$\pm$0.0300/0.4465$\pm$0.0586 with CBraMod and 0.5139$\pm$0.0115/0.5243$\pm$0.0270 with LaBraM; the corresponding \modelnickname{} values are 0.5579$\pm$0.0310 and 0.5671$\pm$0.0032.

\subsection{RGD Isolation and Sensitivity}
\label{app:rgd_controls}

The same-loss controls retain all four training objectives and replace RGD with a parameter-matched MLP or attention factorizer. The attention and RGD variants contain 1,624,444/1,624,835 parameters for CBraMod and 350,868/351,235 for LaBraM.

\begin{table}[!t]
\centering
\small
\caption{Same-loss architecture and recursion-depth controls on FACED. Values are test B-ACC mean$\pm$std.}
\label{tab:app_rgd_isolation}
\setlength{\tabcolsep}{4pt}
\begin{tabular}{@{}lccccc@{}}
\toprule
\textbf{Backbone} & \textbf{MLP} & \textbf{Attention} & \textbf{2 steps} & \textbf{4-step \modelnickname{}} & \textbf{6 steps} \\
\midrule
CBraMod & 0.5794$\pm$0.0045 & 0.5769$\pm$0.0083 & 0.5957$\pm$0.0008 & \bestperf{0.6005$\pm$0.0101} & 0.5932$\pm$0.0003 \\
LaBraM & 0.5081$\pm$0.0098 & 0.5068$\pm$0.0085 & 0.5122$\pm$0.0060 & \bestperf{0.5409$\pm$0.0086} & 0.5208$\pm$0.0058 \\
\bottomrule
\end{tabular}
\end{table}

The attention control obtains 0.4395/0.5208 on BCIC-IV-2a and 0.6017/0.7156 on MentalArith. with CBraMod/LaBraM, compared with \modelnickname{} at 0.5579/0.5671 and 0.7292/0.7236. Fixed gate, fixed precision, static transition, and shared transition reduce FACED B-ACC by 0.0104--0.0305 for CBraMod and 0.0226--0.0404 for LaBraM. Four refinement steps give the highest mean B-ACC in the tested settings.

\begin{table}[!t]
\centering
\small
\caption{Temporal-prior sensitivity on ISRUC-S1 under a fixed multi-seed diagnostic protocol.}
\label{tab:app_temporal_sensitivity}
\setlength{\tabcolsep}{5pt}
\begin{tabular}{@{}lccccc@{}}
\toprule
\textbf{Setting} & \textbf{Default} & \textbf{w/o Invar.} & \textbf{w/o VarRatio} & \textbf{Weight 0.5$\times$} & \textbf{Weight 2$\times$} \\
\midrule
Test B-ACC & \bestperf{0.7922$\pm$0.0015} & 0.7849$\pm$0.0019 & 0.7838$\pm$0.0121 & 0.7809$\pm$0.0133 & 0.7815$\pm$0.0035 \\
\bottomrule
\end{tabular}
\end{table}

\subsection{Objective and Cross-Backbone Controls}
\label{app:objective_combinations}

\begin{table}[!t]
\centering
\small
\caption{Objective-combination controls on CBraMod. Values are test B-ACC under the stated diagnostic protocol.}
\label{tab:app_objective_combinations}
\setlength{\tabcolsep}{5pt}
\begin{tabular}{@{}lccccc@{}}
\toprule
\textbf{Dataset} & \textbf{Vanilla} & \textbf{Variance only} & \textbf{Variance + contrast} & \textbf{Variance + rec.} & \textbf{Full} \\
\midrule
FACED & 0.5387 & 0.5961 & 0.6040 & 0.5805 & \bestperf{0.6126} \\
MentalArith. & 0.6521 & 0.5740 & -- & -- & \bestperf{0.7260} \\
\bottomrule
\end{tabular}
\end{table}

The objective contributions vary by task. Variance-only improves FACED, while the full objective gives the highest result on both datasets.

\begin{table}[!t]
\centering
\small
\caption{LaBraM component controls. Values are test B-ACC.}
\label{tab:app_labram_ablation}
\setlength{\tabcolsep}{6pt}
\begin{tabular}{@{}lcccc@{}}
\toprule
\textbf{Dataset} & \textbf{Full} & \textbf{w/o Invar.} & \textbf{w/o Gaussian} & \textbf{w/o Rec.} \\
\midrule
FACED & \bestperf{0.5480} & 0.5425 & 0.4936 & 0.5157 \\
MentalArith. & \bestperf{0.7417} & 0.6729 & 0.6365 & 0.6635 \\
\bottomrule
\end{tabular}
\end{table}

\subsection{Additional Backbone and Dataset Checks}

REVE \citep{elreve} exposes a post-backbone token sequence compatible with the same adapter interface. Table~\ref{tab:app_reve} reports three subject-disjoint compatibility checks.

\begin{table}[!t]
\centering
\small
\caption{REVE compatibility checks. Values are test B-ACC.}
\label{tab:app_reve}
\setlength{\tabcolsep}{7pt}
\begin{tabular}{@{}lccc@{}}
\toprule
\textbf{Dataset} & \textbf{REVE Vanilla} & \textbf{REVE + \modelnickname{}} & $\boldsymbol{\Delta}$ \\
\midrule
MentalArith. & 0.6531 & \bestperf{0.7083} & +0.0552 \\
BCIC-Speech & 0.6215 & \bestperf{0.6393} & +0.0178 \\
BCIC-IV-2a & 0.5712 & \bestperf{0.5773} & +0.0061 \\
\bottomrule
\end{tabular}
\end{table}

We also evaluate a trait-level EEG target and one non-EEG time-series setting. OpenNeuro Depression REST uses dataset ds003478 v1.1.0 \citep{cavanagh_eeg_2021}. The external check uses MOMENT representations \citep{goswami_moment_2024} on UCI HAR \citep{anguita_public_2013} with a subject-disjoint split.

\begin{table}[!t]
\centering
\small
\caption{Additional subject-disjoint checks. Values are mean$\pm$std.}
\label{tab:app_external_checks}
\setlength{\tabcolsep}{4pt}
\begin{tabular}{@{}llccc@{}}
\toprule
\textbf{Setting} & \textbf{Method} & \textbf{B-ACC} & \textbf{ROC-AUC} & \textbf{PR-AUC / added params} \\
\midrule
OpenNeuro REST & Vanilla & 0.4968$\pm$0.0028 & 0.4150$\pm$0.0105 & 0.2712$\pm$0.0035 \\
OpenNeuro REST & \modelnickname{} & \bestperf{0.5439$\pm$0.0278} & \bestperf{0.6623$\pm$0.0137} & \bestperf{0.4276$\pm$0.0227} \\
\midrule
MOMENT/UCI HAR & Vanilla & 0.5665$\pm$0.0078 & -- & 0 \\
MOMENT/UCI HAR & Generic MLP & 0.7027$\pm$0.0134 & -- & 2,965,836 \\
MOMENT/UCI HAR & \modelnickname{} & \bestperf{0.7742$\pm$0.0249} & -- & 2,966,051 \\
\bottomrule
\end{tabular}
\end{table}

These checks cover one additional EFM, one independent MDD dataset, and one pretrained time-series model/benchmark pair.

\subsection{Probe and Sampler Diagnostics}

Frozen subject probes show that the accessibility of subject identity depends on both backbone and dataset.

\begin{table}[!t]
\centering
\small
\caption{Frozen subject-probe B-ACC across backbones.}
\label{tab:app_backbone_subject_probe}
\setlength{\tabcolsep}{6pt}
\begin{tabular}{@{}lccccc@{}}
\toprule
\textbf{Dataset} & \textbf{Raw EEG} & \textbf{CBraMod} & \textbf{CodeBrain} & \textbf{CSBrain} & \textbf{LaBraM} \\
\midrule
FACED & 0.2297 & 0.6381 & 0.7580 & 0.5611 & \bestperf{0.8598} \\
BCIC-Speech & 0.2434 & 0.3450 & 0.2666 & \bestperf{0.3464} & 0.2874 \\
\bottomrule
\end{tabular}
\end{table}

The subject-balanced sampler gives every anchor at least one same-subject positive in the six main datasets. Anchors without positives are excluded from $\mathcal{L}_{\mathrm{con}}$ as defined in Section~\ref{sec:external_constraint}.

\begin{table}[!t]
\centering
\small
\caption{Effective contrastive-anchor coverage and mean positives per anchor.}
\label{tab:app_anchor_coverage}
\setlength{\tabcolsep}{5pt}
\begin{tabular}{@{}lccc@{}}
\toprule
\textbf{Dataset} & \textbf{Random anchors} & \textbf{DEFT sampler} & \textbf{Positives/anchor} \\
\midrule
BCIC-IV-2a & 99.85\% & 100\% & 6.9701 \\
ISRUC-S1 & 7.78\% & 100\% & 1.0736 \\
Mumtaz2016 & 53.17\% & 100\% & 1.7368 \\
FACED & 16.55\% & 100\% & 1.1593 \\
MentalArith. & 74.56\% & 100\% & 2.4592 \\
BCIC-Speech & 99.92\% & 100\% & 7.8036 \\
\bottomrule
\end{tabular}
\end{table}

\subsection{Temporal and Statistical Diagnostics}

Full samples accumulate more task evidence than one-second windows in both datasets, while subject information is already accessible from short windows.

\begin{table}[!t]
\centering
\small
\caption{One-second versus full-sample task and subject probes. All values are B-ACC.}
\label{tab:app_short_window}
\setlength{\tabcolsep}{4pt}
\begin{tabular}{@{}lcccccc@{}}
\toprule
\textbf{Dataset} & \textbf{Task chance} & \textbf{1-s task} & \textbf{Full task} & \textbf{Subject chance} & \textbf{1-s subject} & \textbf{Full subject} \\
\midrule
MentalArith. & 0.5000 & 0.5585 & \bestperf{0.6316} & 0.0280 & 0.2104 & \bestperf{0.2570} \\
BCIC-IV-2a & 0.2500 & 0.3073 & \bestperf{0.4630} & 0.1111 & 0.6406 & \bestperf{0.6550} \\
\bottomrule
\end{tabular}
\end{table}

At the measured checkpoint, $\mathcal{L}_{\mathrm{ord}}$ and $\mathcal{L}_{\mathrm{sty}}$ have nonzero raw gradient norms of 9.9960 and 0.4042. Their weighted RGD-gradient/task-gradient ratios are 0.0167 and $5.41\times10^{-5}$, with task-gradient cosine similarities of 0.2420 and 0.1246.

\begin{table}[!t]
\centering
\small
\caption{Confidence intervals for the reviewer-identified secondary-metric differences.}
\label{app:secondary_metric_uncertainty}
\setlength{\tabcolsep}{5pt}
\begin{tabular}{@{}llcc@{}}
\toprule
\textbf{Setting} & \textbf{Metric} & $\boldsymbol{\Delta}$ & \textbf{Welch 95\% CI} \\
\midrule
CBraMod--ISRUC-S1 & Kappa & -0.0229 & $[-0.0475, 0.0017]$ \\
CBraMod--ISRUC-S1 & F1 & -0.0145 & $[-0.0374, 0.0084]$ \\
CodeBrain--Mumtaz2016 & ROC-AUC & -0.0074 & $[-0.0368, 0.0220]$ \\
CodeBrain--Mumtaz2016 & PR-AUC & -0.0045 & $[-0.0236, 0.0146]$ \\
LaBraM--FACED & Kappa & +0.0093 & $[-0.0114, 0.0300]$ \\
LaBraM--FACED & F1 & +0.0112 & $[-0.0088, 0.0312]$ \\
\bottomrule
\end{tabular}
\end{table}

\subsection{Additional CBraMod Ablation Results}
\label{app:complete_ablation}

Table~\ref{tab:app_cbramod_ablation} retains the seed-0 metric-level CBraMod ablations over the complete evaluation suite in two panels: style-contrast and variance constraints in Panel A, and Gaussian-disentangler and reconstruction ablations in Panel B. Table~\ref{tab:ablation} is the multi-seed B-ACC comparison used for the main conclusions.

\begingroup
\setlength{\LTleft}{-0.025\textwidth}
\setlength{\LTright}{-0.055\textwidth}
\setlength{\LTcapwidth}{1.08\textwidth}
\setlength{\tabcolsep}{2.0pt}
\let\origaddlinespace\addlinespace
\renewcommand{\addlinespace}[1][]{}
\renewcommand{\arraystretch}{0.84}
\begin{longtable}{@{}>{\small\raggedright\arraybackslash}p{0.155\textwidth}>{\small\raggedright\arraybackslash}p{0.102\textwidth}*{7}{>{\small\centering\arraybackslash}p{0.104\textwidth}}@{}}
\caption{Complete seed-0 CBraMod ablations, Panel A: style contrast and variance constraints. Orange/blue cells mark positive/negative deltas; bold marks the best value.}
\label{tab:app_cbramod_ablation}\\
\toprule[1.1pt]
\textbf{Dataset} & \textbf{Metric} & \textbf{Full} & \textbf{\scriptsize w/o StyleCL} & \textbf{$\Delta$} & \textbf{\scriptsize w/o Invar} & \textbf{$\Delta$} & \textbf{\scriptsize w/o VarRatio} & \textbf{$\Delta$} \\
\midrule[0.4pt]
\endfirsthead
\toprule[1.1pt]
\textbf{Dataset} & \textbf{Metric} & \textbf{Full} & \textbf{\scriptsize w/o StyleCL} & \textbf{$\Delta$} & \textbf{\scriptsize w/o Invar} & \textbf{$\Delta$} & \textbf{\scriptsize w/o VarRatio} & \textbf{$\Delta$} \\
\midrule[0.4pt]
\endhead
\bottomrule[0.8pt]
\endfoot
BCIC-IV-2a & B-ACC & \bestperf{0.5981} & 0.4844 & \negdelta{-0.1137} & 0.3403 & \negdelta{-0.2578} & 0.3715 & \negdelta{-0.2266} \\
 & Kappa & \bestperf{0.4641} & 0.3125 & \negdelta{-0.1516} & 0.1204 & \negdelta{-0.3438} & 0.1620 & \negdelta{-0.3021} \\
 & F1 & \bestperf{0.5954} & 0.4673 & \negdelta{-0.1281} & 0.2757 & \negdelta{-0.3197} & 0.3043 & \negdelta{-0.2910} \\
\addlinespace[1pt]
BCIC-Speech & B-ACC & \bestperf{0.5116} & 0.4880 & \negdelta{-0.0236} & 0.4911 & \negdelta{-0.0204} & 0.4778 & \negdelta{-0.0338} \\
 & Kappa & \bestperf{0.2673} & 0.2320 & \negdelta{-0.0353} & 0.2367 & \negdelta{-0.0307} & 0.2167 & \negdelta{-0.0507} \\
 & F1 & \bestperf{0.5124} & 0.4873 & \negdelta{-0.0251} & 0.4910 & \negdelta{-0.0214} & 0.4758 & \negdelta{-0.0366} \\
\addlinespace[1pt]
FACED & B-ACC & \bestperf{0.6126} & 0.5837 & \negdelta{-0.0289} & 0.6005 & \negdelta{-0.0121} & 0.5993 & \negdelta{-0.0133} \\
 & Kappa & \bestperf{0.5605} & 0.5302 & \negdelta{-0.0303} & 0.5451 & \negdelta{-0.0154} & 0.5448 & \negdelta{-0.0157} \\
 & F1 & \bestperf{0.6128} & 0.5949 & \negdelta{-0.0179} & 0.5998 & \negdelta{-0.0130} & 0.6019 & \negdelta{-0.0109} \\
\addlinespace[1pt]
Stieger2021-UD & B-ACC & \bestperf{0.7730} & 0.7653 & \negdelta{-0.0077} & 0.7629 & \negdelta{-0.0101} & 0.7688 & \negdelta{-0.0042} \\
 & \mbox{ROC-AUC} & \bestperf{0.8575} & 0.8566 & \negdelta{-0.0009} & 0.8506 & \negdelta{-0.0069} & 0.8546 & \negdelta{-0.0029} \\
 & \mbox{PR-AUC} & 0.8633 & \bestperf{0.8659} & \posdelta{+0.0026} & 0.8574 & \negdelta{-0.0059} & 0.8637 & \posdelta{+0.0004} \\
\addlinespace[1pt]
MI-KoreaU & B-ACC & \bestperf{0.8450} & 0.8243 & \negdelta{-0.0207} & 0.8114 & \negdelta{-0.0336} & 0.8275 & \negdelta{-0.0175} \\
 & \mbox{ROC-AUC} & \bestperf{0.9170} & 0.9021 & \negdelta{-0.0150} & 0.8991 & \negdelta{-0.0179} & 0.9053 & \negdelta{-0.0117} \\
 & \mbox{PR-AUC} & \bestperf{0.9221} & 0.9067 & \negdelta{-0.0153} & 0.9057 & \negdelta{-0.0163} & 0.9098 & \negdelta{-0.0122} \\
\addlinespace[1pt]
MentalArith. & B-ACC & \bestperf{0.7260} & 0.5990 & \negdelta{-0.1271} & 0.7104 & \negdelta{-0.0156} & 0.6281 & \negdelta{-0.0979} \\
 & \mbox{ROC-AUC} & \bestperf{0.8376} & 0.6326 & \negdelta{-0.2050} & 0.8133 & \negdelta{-0.0243} & 0.7009 & \negdelta{-0.1367} \\
 & \mbox{PR-AUC} & \bestperf{0.6743} & 0.4364 & \negdelta{-0.2379} & 0.6101 & \negdelta{-0.0642} & 0.5255 & \negdelta{-0.1488} \\
\addlinespace[1pt]
Mumtaz2016 & B-ACC & 0.9644 & 0.9058 & \negdelta{-0.0586} & \bestperf{0.9708} & \posdelta{+0.0064} & 0.8999 & \negdelta{-0.0645} \\
 & \mbox{ROC-AUC} & \bestperf{0.9943} & 0.9687 & \negdelta{-0.0256} & 0.9813 & \negdelta{-0.0130} & 0.9687 & \negdelta{-0.0256} \\
 & \mbox{PR-AUC} & \bestperf{0.9958} & 0.9773 & \negdelta{-0.0185} & 0.9865 & \negdelta{-0.0093} & 0.9733 & \negdelta{-0.0225} \\
\addlinespace[1pt]
PhysioNet & B-ACC & \bestperf{0.6424} & 0.6151 & \negdelta{-0.0272} & 0.6346 & \negdelta{-0.0078} & 0.6263 & \negdelta{-0.0161} \\
 & Kappa & \bestperf{0.5230} & 0.4868 & \negdelta{-0.0362} & 0.5127 & \negdelta{-0.0104} & 0.5016 & \negdelta{-0.0214} \\
 & F1 & \bestperf{0.6413} & 0.6156 & \negdelta{-0.0258} & 0.6338 & \negdelta{-0.0075} & 0.6254 & \negdelta{-0.0159} \\
\addlinespace[1pt]
SEED & B-ACC & \bestperf{0.7877} & 0.7395 & \negdelta{-0.0481} & 0.7111 & \negdelta{-0.0765} & 0.7444 & \negdelta{-0.0432} \\
 & \mbox{ROC-AUC} & \bestperf{0.8589} & 0.8390 & \negdelta{-0.0199} & 0.7931 & \negdelta{-0.0658} & 0.8334 & \negdelta{-0.0256} \\
 & \mbox{PR-AUC} & \bestperf{0.8647} & 0.8255 & \negdelta{-0.0392} & 0.7869 & \negdelta{-0.0778} & 0.8140 & \negdelta{-0.0507} \\
\addlinespace[1pt]
SEED-IV & B-ACC & \bestperf{0.3989} & 0.3483 & \negdelta{-0.0505} & 0.3486 & \negdelta{-0.0503} & 0.3583 & \negdelta{-0.0405} \\
 & Kappa & \bestperf{0.1897} & 0.1222 & \negdelta{-0.0675} & 0.1201 & \negdelta{-0.0696} & 0.1347 & \negdelta{-0.0550} \\
 & F1 & \bestperf{0.3444} & 0.2655 & \negdelta{-0.0788} & 0.2629 & \negdelta{-0.0814} & 0.2841 & \negdelta{-0.0603} \\
\addlinespace[1pt]
SEED-V & B-ACC & 0.2734 & \bestperf{0.2742} & \posdelta{+0.0007} & 0.2734 & 0.0000 & 0.2734 & 0.0000 \\
 & Kappa & 0.0970 & \bestperf{0.0977} & \posdelta{+0.0006} & 0.0970 & 0.0000 & 0.0970 & 0.0000 \\
 & F1 & \bestperf{0.2746} & 0.2720 & \negdelta{-0.0027} & \bestperf{0.2746} & 0.0000 & \bestperf{0.2746} & 0.0000 \\
\addlinespace[1pt]
ISRUC-S1 & B-ACC & \bestperf{0.7925} & 0.7914 & \negdelta{-0.0011} & 0.7843 & \negdelta{-0.0082} & 0.7681 & \negdelta{-0.0244} \\
 & Kappa & 0.7423 & \bestperf{0.7603} & \posdelta{+0.0180} & 0.7398 & \negdelta{-0.0026} & 0.7192 & \negdelta{-0.0231} \\
 & F1 & 0.8049 & \bestperf{0.8135} & \posdelta{+0.0086} & 0.7991 & \negdelta{-0.0058} & 0.7787 & \negdelta{-0.0262} \\
\end{longtable}

\begin{center}
{\small\textit{Table~\ref{tab:app_cbramod_ablation} continued, Panel B: Gaussian disentangler and reconstruction ablations.}}
\end{center}
\vspace{0.15em}

\setlength{\LTleft}{0pt plus 1fill}
\setlength{\LTright}{0pt plus 1fill}
\setlength{\tabcolsep}{2.0pt}
\renewcommand{\arraystretch}{0.84}
\begin{longtable}{@{}>{\small\raggedright\arraybackslash}p{0.17\textwidth}>{\small\raggedright\arraybackslash}p{0.10\textwidth}*{5}{>{\small\centering\arraybackslash}p{0.14\textwidth}}@{}}
\toprule[1.1pt]
\textbf{Dataset} & \textbf{Metric} & \textbf{Full} & \textbf{\scriptsize w/o Gaussian} & \textbf{$\Delta$} & \textbf{\scriptsize w/o Rec.} & \textbf{$\Delta$} \\
\midrule[0.4pt]
\endfirsthead
\multicolumn{7}{@{}l}{\textit{Table~\ref{tab:app_cbramod_ablation} continued, Panel B.}}\\
\toprule[1.1pt]
\textbf{Dataset} & \textbf{Metric} & \textbf{Full} & \textbf{\scriptsize w/o Gaussian} & \textbf{$\Delta$} & \textbf{\scriptsize w/o Rec.} & \textbf{$\Delta$} \\
\midrule[0.4pt]
\endhead
\bottomrule[0.8pt]
\endfoot
BCIC-IV-2a & B-ACC & \bestperf{0.5981} & 0.3967 & \negdelta{-0.2014} & 0.5122 & \negdelta{-0.0859} \\
 & Kappa & \bestperf{0.4641} & 0.1956 & \negdelta{-0.2685} & 0.3495 & \negdelta{-0.1146} \\
 & F1 & \bestperf{0.5954} & 0.3560 & \negdelta{-0.2394} & 0.5096 & \negdelta{-0.0858} \\
\addlinespace[1pt]
BCIC-Speech & B-ACC & \bestperf{0.5116} & 0.4920 & \negdelta{-0.0196} & 0.4773 & \negdelta{-0.0342} \\
 & Kappa & \bestperf{0.2673} & 0.2380 & \negdelta{-0.0293} & 0.2160 & \negdelta{-0.0513} \\
 & F1 & \bestperf{0.5124} & 0.4917 & \negdelta{-0.0207} & 0.4757 & \negdelta{-0.0367} \\
\addlinespace[1pt]
FACED & B-ACC & \bestperf{0.6126} & 0.5926 & \negdelta{-0.0200} & 0.6040 & \negdelta{-0.0086} \\
 & Kappa & \bestperf{0.5605} & 0.5384 & \negdelta{-0.0221} & 0.5518 & \negdelta{-0.0087} \\
 & F1 & \bestperf{0.6128} & 0.5985 & \negdelta{-0.0143} & 0.6107 & \negdelta{-0.0021} \\
\addlinespace[1pt]
Stieger2021-UD & B-ACC & \bestperf{0.7730} & 0.7568 & \negdelta{-0.0162} & 0.7563 & \negdelta{-0.0167} \\
 & \mbox{ROC-AUC} & \bestperf{0.8575} & 0.8480 & \negdelta{-0.0095} & 0.8528 & \negdelta{-0.0047} \\
 & \mbox{PR-AUC} & 0.8633 & 0.8524 & \negdelta{-0.0109} & 0.8598 & \negdelta{-0.0035} \\
\addlinespace[1pt]
MI-KoreaU & B-ACC & \bestperf{0.8450} & 0.8207 & \negdelta{-0.0243} & 0.8311 & \negdelta{-0.0139} \\
 & \mbox{ROC-AUC} & \bestperf{0.9170} & 0.9013 & \negdelta{-0.0157} & 0.9138 & \negdelta{-0.0032} \\
 & \mbox{PR-AUC} & \bestperf{0.9221} & 0.9083 & \negdelta{-0.0137} & 0.9200 & \negdelta{-0.0020} \\
\addlinespace[1pt]
MentalArith. & B-ACC & \bestperf{0.7260} & 0.5521 & \negdelta{-0.1740} & 0.6531 & \negdelta{-0.0729} \\
 & \mbox{ROC-AUC} & \bestperf{0.8376} & 0.7216 & \negdelta{-0.1160} & 0.7507 & \negdelta{-0.0869} \\
 & \mbox{PR-AUC} & \bestperf{0.6743} & 0.5250 & \negdelta{-0.1493} & 0.5037 & \negdelta{-0.1707} \\
\addlinespace[1pt]
Mumtaz2016 & B-ACC & 0.9644 & 0.9272 & \negdelta{-0.0372} & 0.8997 & \negdelta{-0.0647} \\
 & \mbox{ROC-AUC} & \bestperf{0.9943} & 0.9040 & \negdelta{-0.0903} & 0.9013 & \negdelta{-0.0930} \\
 & \mbox{PR-AUC} & \bestperf{0.9958} & 0.9459 & \negdelta{-0.0499} & 0.9463 & \negdelta{-0.0495} \\
\addlinespace[1pt]
PhysioNet & B-ACC & \bestperf{0.6424} & 0.6152 & \negdelta{-0.0272} & 0.6335 & \negdelta{-0.0089} \\
 & Kappa & \bestperf{0.5230} & 0.4868 & \negdelta{-0.0362} & 0.5112 & \negdelta{-0.0118} \\
 & F1 & \bestperf{0.6413} & 0.6145 & \negdelta{-0.0268} & 0.6322 & \negdelta{-0.0092} \\
\addlinespace[1pt]
SEED & B-ACC & \bestperf{0.7877} & 0.7481 & \negdelta{-0.0395} & 0.7395 & \negdelta{-0.0481} \\
 & \mbox{ROC-AUC} & \bestperf{0.8589} & 0.8407 & \negdelta{-0.0182} & 0.8101 & \negdelta{-0.0488} \\
 & \mbox{PR-AUC} & \bestperf{0.8647} & 0.8228 & \negdelta{-0.0419} & 0.7849 & \negdelta{-0.0798} \\
\addlinespace[1pt]
SEED-IV & B-ACC & \bestperf{0.3989} & 0.3752 & \negdelta{-0.0237} & 0.3152 & \negdelta{-0.0836} \\
 & Kappa & \bestperf{0.1897} & 0.1564 & \negdelta{-0.0333} & 0.0774 & \negdelta{-0.1123} \\
 & F1 & \bestperf{0.3444} & 0.3057 & \negdelta{-0.0387} & 0.2161 & \negdelta{-0.1282} \\
\addlinespace[1pt]
SEED-V & B-ACC & 0.2734 & 0.2634 & \negdelta{-0.0100} & 0.2714 & \negdelta{-0.0020} \\
 & Kappa & 0.0970 & 0.0831 & \negdelta{-0.0139} & 0.0937 & \negdelta{-0.0033} \\
 & F1 & \bestperf{0.2746} & 0.2659 & \negdelta{-0.0087} & 0.2700 & \negdelta{-0.0046} \\
\addlinespace[1pt]
ISRUC-S1 & B-ACC & \bestperf{0.7925} & 0.7782 & \negdelta{-0.0143} & 0.7801 & \negdelta{-0.0124} \\
 & Kappa & 0.7423 & 0.7395 & \negdelta{-0.0029} & 0.7279 & \negdelta{-0.0145} \\
 & F1 & 0.8049 & 0.7974 & \negdelta{-0.0075} & 0.7889 & \negdelta{-0.0160} \\
\end{longtable}
\endgroup

The seed-0 records provide metric-level detail, while the fixed-seed comparison in Table~\ref{tab:ablation} supports the task-dependent component analysis in the main text.


\newpage
\section*{NeurIPS Paper Checklist}

\begin{enumerate}

\item {\bf Claims}
    \item[] Question: Do the main claims made in the abstract and introduction accurately reflect the paper's contributions and scope?
    \item[] Answer: \answerYes{}.
    \item[] Justification: The abstract and introduction state the contribution as a post-backbone disentanglement adapter for EFM fine-tuning. The main quantitative scope is defined in Section~\ref{sec:main_result}, with limitations discussed in Appendix~\ref{app:limitations_ethics_reproducibility}.

\item {\bf Limitations}
    \item[] Question: Does the paper discuss the limitations of the work performed by the authors?
    \item[] Answer: \answerYes{}.
    \item[] Justification: Appendix~\ref{app:limitations_ethics_reproducibility} discusses the use of validation-selected loss weights, the temporal-variance prior, training-time subject identities, and the remaining direction of fully unsupervised disentanglement fine-tuning.

\item {\bf Theory assumptions and proofs}
    \item[] Question: For each theoretical result, does the paper provide the full set of assumptions and a complete (and correct) proof?
    \item[] Answer: \answerNA{}.
    \item[] Justification: The paper does not present formal theoretical results. Method derivations used to specify the Gaussian update are provided in Appendix~\ref{app:gaussian_details}.

    \item {\bf Experimental result reproducibility}
    \item[] Question: Does the paper fully disclose all the information needed to reproduce the main experimental results of the paper to the extent that it affects the main claims and/or conclusions of the paper (regardless of whether the code and data are provided or not)?
    \item[] Answer: \answerYes{}.
    \item[] Justification: Sections~\ref{sec:exp_setup}--\ref{sec:main_result} define the evaluation setup, datasets, backbones, metrics, and main comparison. Appendices~\ref{app:dataset_details}--\ref{app:implementation_details} provide split statistics, preprocessing, hyperparameters, and implementation details, and Appendix~\ref{app:limitations_ethics_reproducibility} provides the code repository link.

\item {\bf Open access to data and code}
    \item[] Question: Does the paper provide open access to the data and code, with sufficient instructions to faithfully reproduce the main experimental results, as described in supplemental material?
    \item[] Answer: \answerYes{}.
    \item[] Justification: The experiments use released EEG datasets, and code with configuration files is provided at the link in Appendix~\ref{app:limitations_ethics_reproducibility}.

\item {\bf Experimental setting/details}
    \item[] Question: Does the paper specify all the training and test details (e.g., data splits, hyperparameters, how they were chosen, type of optimizer) necessary to understand the results?
    \item[] Answer: \answerYes{}.
    \item[] Justification: Section~\ref{sec:exp_setup} summarizes the datasets, backbones, update policy, hardware, and metrics. Appendix~\ref{app:dataset_details} reports split statistics, and Appendix~\ref{app:training_hparams} reports optimizer and hyperparameter policies.

\item {\bf Experiment statistical significance}
    \item[] Question: Does the paper report error bars suitably and correctly defined or other appropriate information about the statistical significance of the experiments?
    \item[] Answer: \answerYes{}.
    \item[] Justification: Table~\ref{tab:main_result} reports mean and standard deviation over available seeds for the main comparison, and Section~\ref{sec:exp_setup} states that the main table uses multi-seed mean and standard deviation.

\item {\bf Experiments compute resources}
    \item[] Question: For each experiment, does the paper provide sufficient information on the computer resources (type of compute workers, memory, time of execution) needed to reproduce the experiments?
    \item[] Answer: \answerYes{}.
    \item[] Justification: Section~\ref{sec:exp_setup} reports the GPU type, Appendix~\ref{app:compute_resources} reports GPU-hour accounting for reproducing the main evaluation and appendix diagnostics, and Appendix~\ref{app:runtime_memory_overhead} reports representative parameter, memory, and one-step FLOP overhead.
    
\item {\bf Code of ethics}
    \item[] Question: Does the research conducted in the paper conform, in every respect, with the NeurIPS Code of Ethics \url{https://neurips.cc/public/EthicsGuidelines}?
    \item[] Answer: \answerYes{}.
    \item[] Justification: The work uses released datasets, subject-disjoint evaluation and training-time subject IDs only for regularization. Ethical and privacy considerations are discussed in Appendix~\ref{app:limitations_ethics_reproducibility}.

\item {\bf Broader impacts}
    \item[] Question: Does the paper discuss both potential positive societal impacts and negative societal impacts of the work performed?
    \item[] Answer: \answerYes{}.
    \item[] Justification: Appendix~\ref{app:limitations_ethics_reproducibility} discusses the research value of more robust cross-subject EEG representation learning and the risk of inappropriate clinical interpretation or privacy-sensitive use of EEG/MDD-related data.
    
\item {\bf Safeguards}
    \item[] Question: Does the paper describe safeguards that have been put in place for responsible release of data or models that have a high risk for misuse (e.g., pre-trained language models, image generators, or scraped datasets)?
    \item[] Answer: \answerNA{}.
    \item[] Justification: The paper does not release a high-risk generative model, scraped dataset, or new sensitive dataset. It releases research code and uses already released EEG datasets.

\item {\bf Licenses for existing assets}
    \item[] Question: Are the creators or original owners of assets (e.g., code, data, models), used in the paper, properly credited and are the license and terms of use explicitly mentioned and properly respected?
    \item[] Answer: \answerYes{}.
    \item[] Justification: The paper cites the original datasets and EFM backbones used in the experiments, and Appendix~\ref{app:asset_licenses} summarizes dataset sources, license names or usage terms, and cases where no standard license name could be identified. Raw datasets and pretrained checkpoints are not redistributed in the code release.

\item {\bf New assets}
    \item[] Question: Are new assets introduced in the paper well documented and is the documentation provided alongside the assets?
    \item[] Answer: \answerYes{}.
    \item[] Justification: The new released asset is the implementation and configuration files, referenced in Appendix~\ref{app:limitations_ethics_reproducibility}. No new dataset is introduced.

\item {\bf Crowdsourcing and research with human subjects}
    \item[] Question: For crowdsourcing experiments and research with human subjects, does the paper include the full text of instructions given to participants and screenshots, if applicable, as well as details about compensation (if any)? 
    \item[] Answer: \answerNA{}.
    \item[] Justification: This work does not conduct new crowdsourcing or collect new human-subject data; it evaluates on previously released EEG datasets.

\item {\bf Institutional review board (IRB) approvals or equivalent for research with human subjects}
    \item[] Question: Does the paper describe potential risks incurred by study participants, whether such risks were disclosed to the subjects, and whether Institutional Review Board (IRB) approvals (or an equivalent approval/review based on the requirements of your country or institution) were obtained?
    \item[] Answer: \answerNA{}.
    \item[] Justification: The paper performs secondary analysis of released EEG datasets and does not collect new data from human participants. The manuscript cites the original dataset sources and reports no new IRB approval for newly collected data.

\item {\bf Declaration of LLM usage}
    \item[] Question: Does the paper describe the usage of LLMs if it is an important, original, or non-standard component of the core methods in this research? Note that if the LLM is used only for writing, editing, or formatting purposes and does \emph{not} impact the core methodology, scientific rigor, or originality of the research, declaration is not required.
    \item[] Answer: \answerNA{}.
    \item[] Justification: The proposed method and experiments do not use LLMs as an important, original, or non-standard component. Any writing or formatting assistance is outside the core methodology and does not require declaration under the checklist wording.

\end{enumerate}

\end{document}